\documentclass[preprint,review,12pt,authoryear]{elsarticle}

\usepackage{amssymb}
\usepackage{amsmath}
\usepackage{xcolor}
\usepackage{soul}
\usepackage{subcaption}
\usepackage{hyperref}
\usepackage{booktabs}
\usepackage{multirow}
\usepackage{threeparttable}
\soulregister{\cite}{7}

\journal{}

\newdefinition{rmk}{Remark}

\begin{document}

\begin{frontmatter}

\title{eCNNTO: A Highly Generalizable ConvNet for Accelerating Topology Optimization} 

\author[1]{Shengbiao Lu} 
\affiliation[1]{organization={Global college, Shanghai Jiao Tong University},
            addressline={800 Dongchuan Road}, 
            city={Shanghai},
            postcode={200240}, 
            state={Shanghai},
            country={China}}

\author[1]{Xiaodong Wei\corref{mycorrespondingauthor}}
\cortext[mycorrespondingauthor]{Corresponding author}
\ead{xiaodong.wei@sjtu.edu.cn}

\begin{abstract}
This work proposes an element-based Convolutional Neural Network (CNN) to accelerate density-based Topology Optimization (TO) , termed \textit{eCNNTO}. TO generally undergoes a large number of iterations, where finite element analysis is performed in every iteration, leading to the efficiency bottleneck especially when dense meshes are used to achieve high-resolution designs. To address this limitation, eCNNTO is proposed to build upon \cite{kalliorasAcceleratedTopologyOptimization2020}, where a Deep Belief Network (DBN) was trained for every element to predict its near-optimal density from its early history, thereby skipping the great majority of iterations and significantly accelerating the TO procedure. However, the method lacks spatial correlations among neighboring elements and may lead to disconnected features in the final structure. The proposed method employs CNN with residual connections to address this issue. On top of it, a novel training strategy is introduced to further enhance the optimization efficiency, where the training dataset consists of the final stage density histories rather than early ones. This change can also help reduce the required training data size. eCNNTO requires only a small dataset to train and yet it can be generalized to problems with largely different boundary conditions, loading cases, design domain geometries, mesh resolutions, as well as non-design domains. In the end, the generalization capabilities and efficiency of eCNNTO are demonstrated through a variety of examples in two and three dimensions, achieving up to 90\% and 97\% reduction of iterations, respectively.

\end{abstract}

%
%
%
%

\begin{keyword}

Strong generalization \sep Topology optimization acceleration \sep CNN \sep Small data

\end{keyword}

\end{frontmatter}

\section{Introduction}\label{sec1}

Topology Optimization (TO) is a structural design methodology that determines the optimal material distribution within a prescribed design domain to achieve the best structural performance under given boundary conditions and constraints. Due to its high design flexibility and ability to generate high-performance structures, topology optimization has been widely applied in various fields such as aerospace engineering \citep{mekkiGeneticAlgorithmBased2021}, bioengineering \citep{xueDesignSelfExpandingAuxetic2020, ahadiTopologyOptimizationCoronary2024}, and automotive design \citep{zhangIntegratedMultiobjectiveTopology2021, suTopologyOptimizationLightweight2022}.

A variety of numerical methods of TO have been developed, such as Solid Isotropic Material with Penalization (SIMP) \citep{bendsoeOptimalShapeDesign1989,andreassenEfficientTopologyOptimization2011,wangEfficientLargescale3D2025}, Evolutionary Structural Optimization (ESO) \citep{xieSimpleEvolutionaryProcedure1993}, Bi-directional Evolutionary Structural Optimization (BESO) \citep{xiaBidirectionalEvolutionaryStructural2018}, Level-Set Methods (LSM) \citep{wangLevelSetMethod2003}, Moving Morphable Components (MMC) \citep{guoDoingTopologyOptimization2014} and phase-field-based methods \citep{wangPhasefieldMethodCombined2024, shengIsogeometricTopologyOptimization2025}. Among these approaches, SIMP has been extensively studied and widely adopted because of its mathematical tractability, implementation simplicity, and practical robustness. However, SIMP is computationally expensive because it requires dense meshes to achieve high-resolution designs, which significantly increase computational cost due to repeated finite element analysis throughout the optimization. It becomes particularly prohibitive for large-scale problems \citep{aageGigavoxelComputationalMorphogenesis2017}.

With the rapid development of artificial intelligence, neural network–based approaches have been increasingly explored to accelerate topology optimization. Existing learning-based methods can be broadly classified into global methods and local methods. Global methods aim to directly generate final topologies from problem settings (e.g., boundary conditions and loading cases), essentially treating the problem as an image-to-image mapping task. Early works by Sosnovik and Oseledets \citep{sosnovikNeuralNetworksTopology2019a} pioneered this approach using 2D Convolutional Neural Networks (CNNs), where a convolutional encoder-decoder architecture to map intermediate densities and their gradients to final structures. \cite{nieTopologyGANTopologyOptimization2021a} adopted Generative Adversarial Networks (GAN) to generate final structures using loading and boundary conditions as the input. \cite{behzadiGANTLPracticalRealTime2021} replaced the network with the conditional GAN (cGAN) to improve the generalization capability to unseen boundary and external loading conditions. Rather than directly generating optimized structures, \cite{qianAcceleratingGradientbasedTopology2021} proposed a dual-network surrogate for forward and sensitivity analyses to accelerate the optimization process. \cite{xingOnlineAutonomousLearning2023} introduced an autonomous online learning strategy, where a neural network is trained on-the-fly using the data collected from the executed SIMP iterations, and then the trained surrogate replaces conventional sensitivity analysis to accelerate the optimization. Though these approaches can significantly reduce the optimization time, they often require large datasets and exhibit limited robustness \citep{woldsethUseArtificialNeural2022, banga3DTopologyOptimization2018b, sosnovikNeuralNetworksTopology2019a}, where small variations in input conditions can lead to disconnected structures or violations of physical constraints. 

To address the robustness issue, several neural nets have been proposed to predict a near-optimal initial design, followed by a correction stage via the conventional SIMP. \cite{padhiDeepLearningAccelerated2024} proposed the conditional invertible neural network to predict near-optimal structures with a few SIMP iterations as the input, which can reduce 40\% of iterations. \cite{limAcceleratingTopologyOptimization2024} proposed a CNN to accelerate SIMP-based optimization and generate high-resolution near-optimal structures. \cite{jooDynamicGraphbasedConvergence2024} proposed a dynamic graph-based neural network to accelerate the convergence of topology optimization for unstructured meshes. Despite their demonstrated efficiency, the practical application of these global methods is hindered by the prohibitive cost of preparing training datasets and their limited generalization capabilities \citep{woldsethUseArtificialNeural2022}. 

In contrast, local methods focus on learning the evolution of element-wise densities. As structural variations follow relatively regular patterns at the local level, these methods typically require fewer training samples and demonstrate improved generalization capabilities. \cite{kalliorasAcceleratedTopologyOptimization2020} proposed the Deep Learning Assisted Topology OPtimization (DLTOP), which uses the early history of element density to predict the near-optimal density, followed by a few SIMP iterations to ensure structural connectivity and physical validity. While DLTOP demonstrated a significant acceleration (more than 50\% reduction of iterations), it predicts each element independently and neglects the spatial correlations among neighboring elements. This lack of spatial context may lead to physically defective features, such as corner contacts and disconnections. To address this issue, \cite{jooUnitModuleBasedConvergence2021} incorporated the spatio-temporal information using Convolutional Long Short-Term Memory (ConvLSTM) networks through a so-called unit module, which is a patch of $64 \times 64$ elements. Their training strategy relied on optimizing unit modules under varying loading conditions, which requires high computational cost to prepare the training datasets and increases the model complexity. 

To address the limitations of existing local approaches, this work, termed \textit{eCNNTO}, builds on top of DLTOP \citep{kalliorasAcceleratedTopologyOptimization2020} and proposes a CNN-based framework for accelerating SIMP-based topology optimization. CNNs are used to explicitly account for spatial correlations among neighboring elements. They are naturally suited for this task as their convolution operations restore the continuum assumption of structures and the filtering scheme used in SIMP. By exploiting spatial correlations, eCNNTO can suppress the occurrence of defective features, such as isolated pieces, corner-contact parts and intermediate-density regions. Furthermore, a novel training strategy is proposed, where the training datasets are prepared using element densities from the \textit{final stage}, rather than those from the early stage in DLTOP. This way, further speedup can be achieved and the size of required training data can be reduced. Moreover, eCNNTO shows strong generalization capabilities. It can accommodate varying boundary conditions, loading cases, design domain geometries, mesh resolutions, and non-design domains.

The rest of the paper is organized as follows: Topology optimization based on SIMP is introduced in Section \ref{Sec2}. eCNNTO is presented in \autoref{Sec3} in detail, including the network architecture, dataset construction and the novel training strategy. Section \ref{Sec4} presents a variety of numerical examples to demonstrate the advantages of eCNNTO. Finally, Section \ref{Sec5} draws conclusions and suggests directions for future research.

\section{Topology Optimization and Deep Learning Assisted Acceleration}\label{Sec2}

\subsection{Solid Isotropic Material with Penalization (SIMP)}\label{Sec21}

In a typical Topology Optimization (TO) problem, an initial material distribution is assumed and then iteratively updated via a certain gradient-based optimization method, which relies on the state of the problem. Such state is obtained by solving certain governing equations through Finite Element Analysis (FEA). An optimized structure is obtained once a certain convergence criterion is satisfied. 

Solid Isotropic Material with Penalization (SIMP) has been widely adopted in TO. It adopts a continuous density between 0 (indicating void) and 1 (pure material) to represent the material distribution, where intermediate values are penalized towards the two extremes (i.e., 0 and 1). The prescribed domain where material can be possibly distributed is called a \textit{design domain}. It is discretized into a mesh, where every element is assigned a density value $\rho_{e} \in [0,1]$. Central to SIMP is the rescaling of the key material parameters (e.g., Young's modulus $E_{e}$) using $\rho_{e}$. Specifically, $E_{e}$ is defined through the penalization strategy:

\begin{equation}\label{Eq:penalization strategy}
    E_{e}(\rho_{e}) = E_\text{min} + \rho_{e}^{p} (E_\text{max} - E_\text{min}),
\end{equation}

\noindent where $E_\text{max}$ is the Young's modulus of pure material, $E_\text{min}$ is a threshold Young's modulus to avoid singular stiffness matrices, and $p$ is a penalization factor (usually $p=3$) used to penalize the behavior of intermediate density values. 

In this work, we use the classical minimum compliance as the model problem to introduce the proposed method. It is formulated as follows:

\begin{equation}\label{Eq:optimizationmodel}
\begin{aligned}
\min_{\mathbf{\rho}} :& \ C(\mathbf{\rho}) = \mathbf{U(\rho)}^\mathrm{T}\mathbf{K(\rho)}\mathbf{U(\rho)}, \\
\text{s.t} : & \quad \mathbf{K(\rho)}\mathbf{U(\rho)} = \mathbf{F}, \\
& \frac{V(\mathbf{\rho})}{V_0} = V_{f}, \\
& 0 \leq \rho_{e} \leq 1, \quad e=1, \ldots, N,
\end{aligned}
\end{equation}

\noindent where $C$ denotes the structural compliance, $\mathbf{U}$, $\mathbf{K}$ and $\mathbf{F}$ are the global displacement vector, the stiffness matrix, and the global force vector, respectively, $V(\rho)$ is the volume of the target structure, $V_{0}$ is the volume of the design domain, $V_{f}$ is the prescribed volume fraction, and $N$ is the number of elements. Note that $\mathbf{U}$, $\mathbf{K}$ and $\mathbf{F}$ are a result of applying FEA to solve the linear elasticity problem.

In \autoref{Eq:optimizationmodel}, element densities $\rho_{e}$ are the \textit{design variables}. Their final values lead to the optimized structure. The optimization process is typically initialized with a constant density field corresponding to $V_{f}$. Given the current $\rho_{e}$, $\mathbf{K}\mathbf{U} = \mathbf{F}$ is solved to find the current state of the problem (i.e., $\mathbf{U}$), with which the compliance and its gradients with respect to $\rho_{e}$ can be evaluated. The constrained optimization problem is then solved using gradient-based optimization algorithms, such as the Optimality Criteria (OC) \citep{sigmund99LineTopology2001a,bendsoeTopologyOptimization2004} and the Method of Moving Asymptotes (MMA) \citep{svanbergMethodMovingAsymptotes1987}. The optimization procedure terminates when a stopping criterion is met. For example, the maximum change of element densities reaches a given threshold $\epsilon$, 

\begin{equation}\label{Eq:convergence criterion}
    \max_{e \in \{1, \ldots, N\}} {|\Delta{\rho_{e}}|} \leq \epsilon.
\end{equation}

 SIMP-based TO often suffers from numerical instabilities such as the checkerboard phenomenon \citep{sigmundNumericalInstabilitiesTopology1998}, where void and solid elements appear alternately. Various filters have been proposed to resolve this issue such as the density filter \citep{sigmundMorphologybasedBlackWhite2007}, the sensitivity filter \citep{sigmundSensitivityFilteringContinuum2012} and the PDE filter \citep{kawamotoHeavisideProjectionBased2011}.

In this work, SIMP iterations are performed using the popular MATLAB codes \cite{andreassenEfficientTopologyOptimization2011,wangEfficientLargescale3D2025}. They are used to prepare our datasets and also verify our method.

\subsection{Deep Learning Assisted Topology Optimization (DLTOP)}\label{Sec22}

\begin{figure*}
\centering
\includegraphics[width=0.80\textwidth]{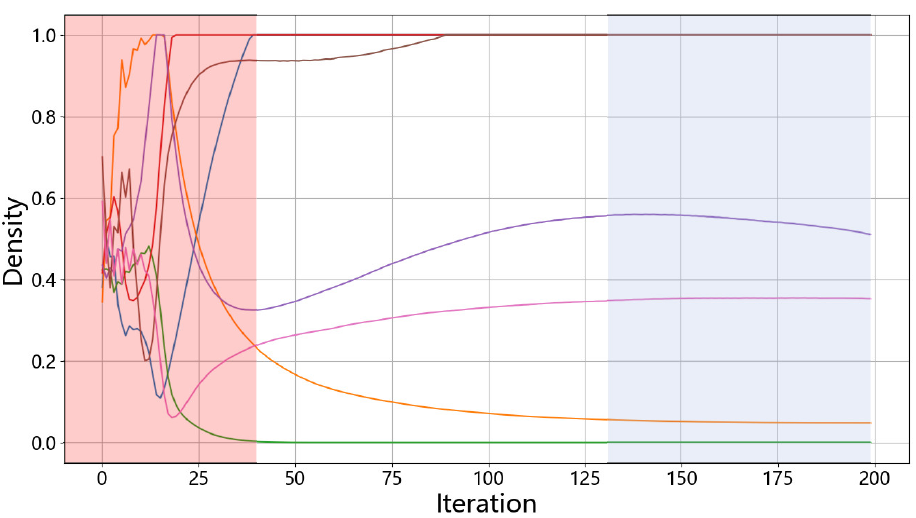}
\caption{Evolution of element densities with respect to the number of iterations. In the early stage (red region), element densities undergo drastic and fluctuating changes, while in the late stage (blue region), density variations tend to be stable.}\label{fig:Ideaexplanation}
\end{figure*}

 Looking at the density evolutions of individual elements (e.g., \autoref{fig:Ideaexplanation}), we find that significant changes mainly occur during the early stage of the optimization, whereas density evolution becomes relatively stable in the later stage. Motivated by this characteristic, an acceleration method named Deep Learning Assisted Topology Optimization (DLTOP) \citep{kalliorasAcceleratedTopologyOptimization2020} was proposed to predict the near-optimal density of an element based on its early history of the density sequence, thereby skipping a large number of intermediate iterations. 
 
 The training dataset of DLTOP is prepared at the element level. Therefore, running a single TO problem can already generate many data samples, and thus data preparation in DLTOP is efficient. A dataset consists of pairs of element density sequences at the early stage and the optimized element densities. Specifically, for every element, the density sequence obtained from the first N iterations serves as the input, whereas its final density is the label. 
 
 Once trained, the neural net can be used for new scenarios without retraining. The input is obtained by performing SIMP for N iterations. Based on this, the neural net predicts a density value for each element, which collectively yields a near-optimal structure. Finally, another few SIMP iterations are performed to improve structural connectivity and satisfy the prescribed volume constraint until the convergence criterion is satisfied. 

It is worth mentioning that DLTOP divides the labels into 3 or 12 classes according to the ranges of density values. Classifications are done by

\begin{equation}\label{Eq:classification criterion1}
    \rho_{i} = \begin{cases}
    0, &\rho_{i} \in [0,0.4],\\
    0.5, &\rho_{i} \in (0.4, 0.7),\\
    1, &\rho_{i} \in [0.7,1],\\
\end{cases}
\end{equation}

\noindent or

\begin{equation}\label{Eq:classification criterion2}
    \rho_{i} = 
    \begin{cases}
        0, & \rho_{i} = 0,\\
        0.05, & \rho _{i}\in (0,0.1],\\
        \vdots & \vdots \\
        0.95, & \rho_{i} \in (0.9,1.0),\\
        1, & \rho_{i} = 1,
    \end{cases}
\end{equation}

\noindent where $\rho_{i}$ ($i=1,2,\ldots$) is a labeled density. Note that only the final densities are classified into discrete values, whereas the inputs remain continuous.

However, DLTOP overlooked the fact that evolutions of neighboring elements depend on one another, leading to the need of relatively large dataset, poor structural connectivity, and even isolated structural features. 

\begin{figure*}
\centering
\includegraphics[width=0.95\textwidth]{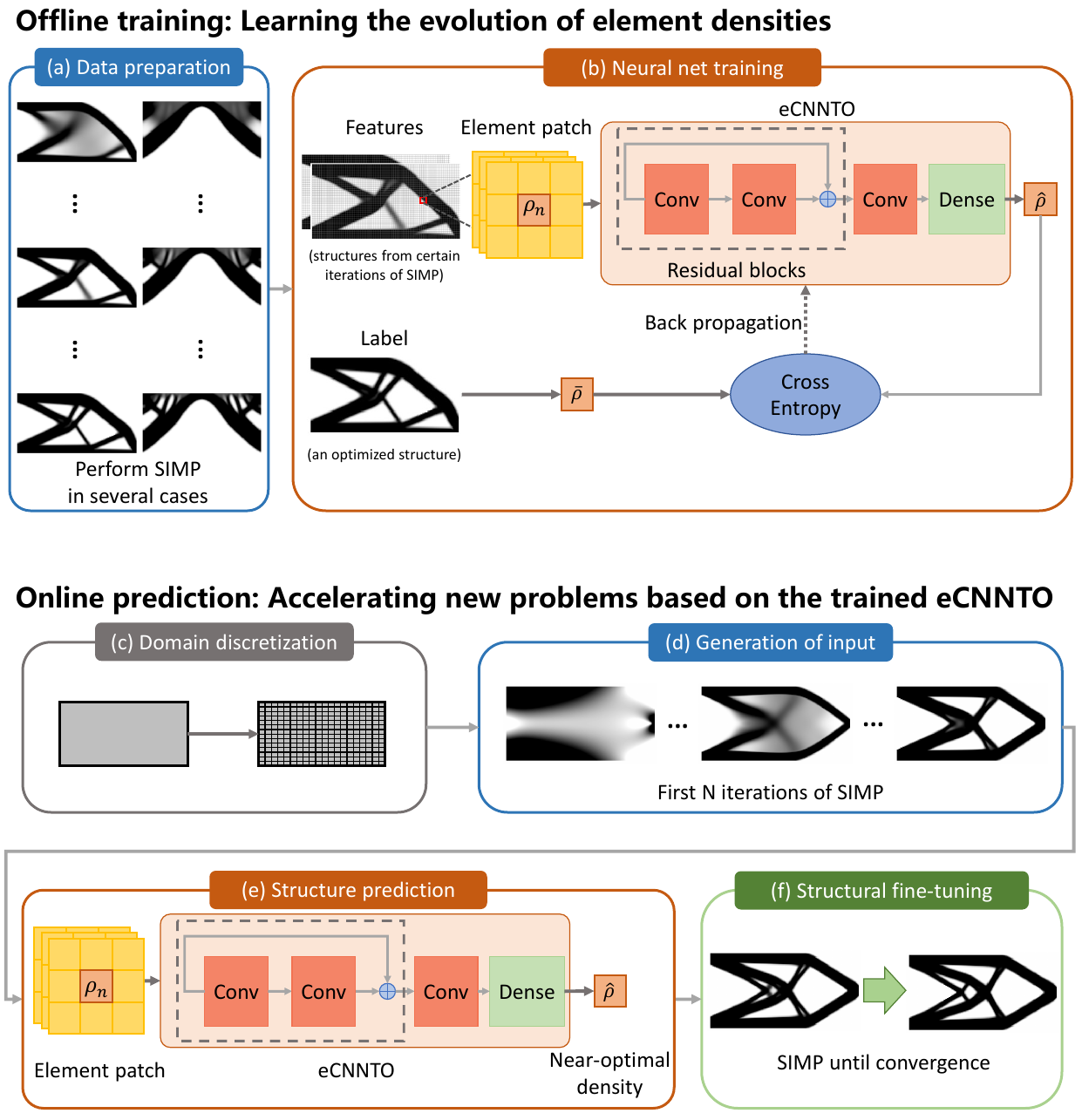}
\caption{Overall workflow of eCNNTO. (a) Preparation of the training dataset, where SIMP is performed under several different settings. (b) eCNNTO is trained on the labeled data at the element level. (c) A design domain is discretized into a mesh with certain resolution. (d) Preparation of the input data, where SIMP is performed only a few iterations. (e) eCNNTO predicts a near-optimal structure and it skips a large number of intermediate iterations needed in SIMP. (f) Structural fine-tuning via SIMP.}\label{fig:Overview framework}
\end{figure*}

\section{Element-Based CNN for Accelerating TO (eCNNTO)}\label{Sec3}

In this section, we introduce the proposed method, namely an element-based Convolutional Neural Network for accelerating Topology Optimization (eCNNTO). \autoref{fig:Overview framework} illustrates the overall workflow of eCNNTO. It is divided into offline training and online prediction. During offline training, eCNNTO learns the evolution of element densities based on the data obtained from SIMP. At the online stage, the trained eCNNTO predicts the near-optimal structures of new problems according to the early histories of element densities (computed by SIMP). Finally, the predicted structures are further optimized by SIMP until convergence. In what follows, we first introduce the architecture of eCNNTO, where CNN and the residual connections \citep{heDeepResidualLearning2016} are used. Next, we discuss the construction of dataset in detail. In the end, a new training strategy is presented to reduce the size of the training dataset through a particular selection of input features.

\subsection{Network architecture}\label{Sec31}

\begin{figure*}
\centering
\includegraphics[width=1.0\textwidth]{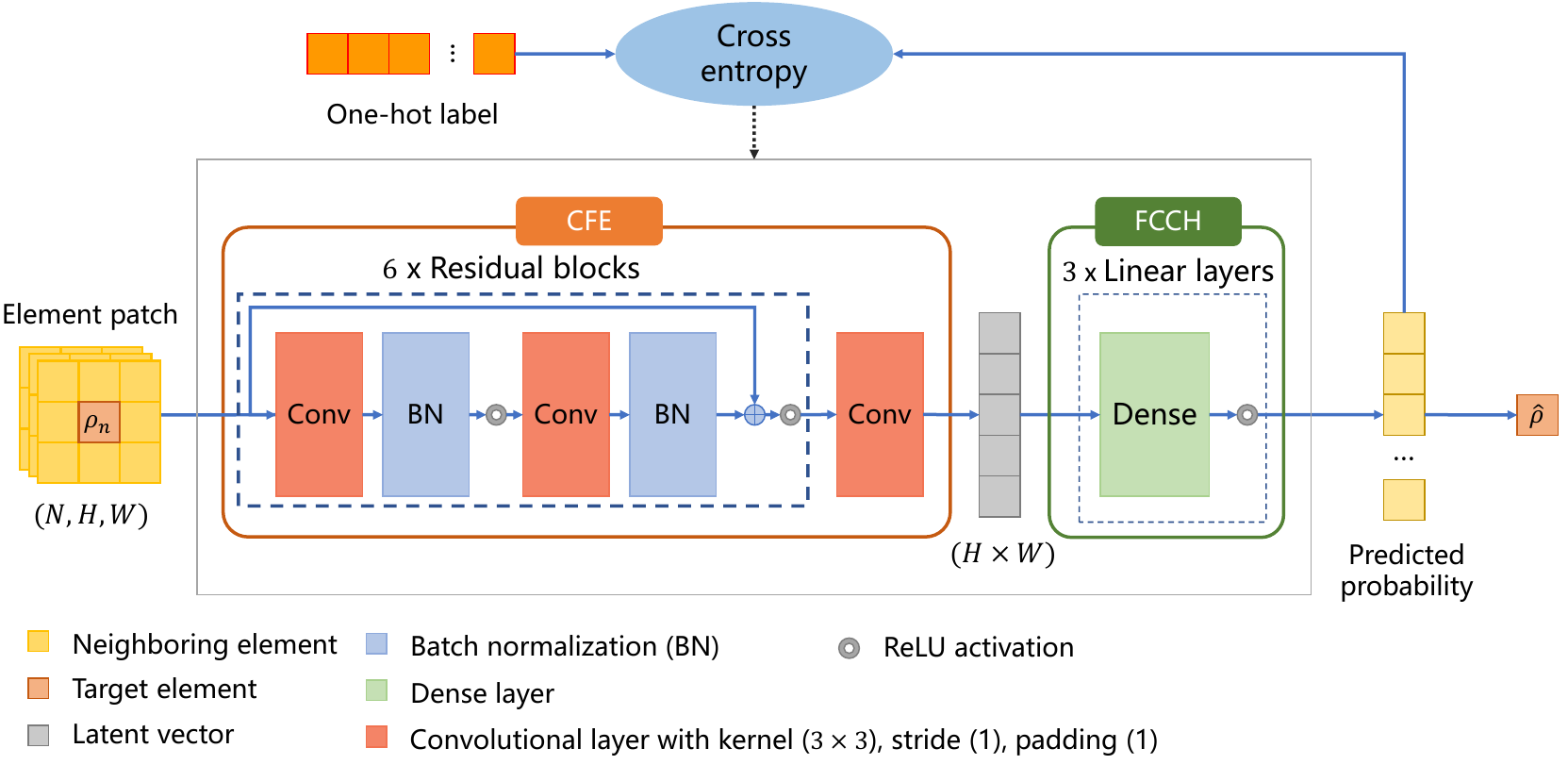}
\caption{Network architecture of eCNNTO. Convolutional Feature Extractor (CFE) contains 6 residual blocks to extract spatial features from the input, which is a density sequence of an element patch. In each block, residual connection is constructed through a shortcut path. Fully Connected Classification Head (FCCH), composed of 3 linear layers, integrates the extracted features from CFE to output the near-optimal density of the target element.}\label{fig:network architecture}
\end{figure*}

We start with the network architecture of eCNNTO; see \autoref{fig:network architecture}. In addition to the input and output, it has two main building blocks: a Convolutional Feature Extractor (CFE) and a Fully Connected Classification Head (FCCH). In the input layer, eCNNTO works with the so-called \textit{element patch}, which consists of the target element and its neighboring elements, together forming a local window of size $H \times W$. The input to the network is a density sequence of an element patch, which is obtained by performing SIMP for N iterations. Thus, the input is a $N \times H \times W$ tensor. In practice, we usually take $H = W = 3$ (3D) or $5$ (2D). Their choice (also $N$) and influence will be further discussed in \autoref{Sec4}.

The element patch is introduced due to the adoption of Convolutional Neural Networks (CNNs) in the proposed architecture. CNNs are a class of deep learning models that leverage local receptive fields and weight sharing to efficiently extract spatial features from grid-based data \citep{lecunDeepLearning2015}. The explicit inductive bias of CNN is necessary for capturing local spatial correlations, making it an ideal candidate to address the poor connectivity issue of DLTOP. 

The input is fed into CFE, which is a typical network architecture used to extract local features. CFE mainly stacks a certain number of residual blocks. Each block features a convolutional layer (Conv), followd by a batch normalization layer (BN) and a ReLU activation function. In a convolutional layer, let $\mathbf{X}^{(l-1)} \in \mathbb{R}^{C_{in} \times H \times W}$ and $\mathbf{W}^{(l)} \in \mathbb{R}^{C_{out} \times C_{in} \times K_H \times K_W}$ be the input feature map and the learnable kernel weights of the $l^{th}$ layer, respectively, where $C_{in}$ and $C_{out}$ are the number of channels of the input and output, and $K_{H}$ and $K_{W}$ are the height and width of the kernel. When $l=1$, $\mathbf{X}^{(0)}$ is simply the input and thus $C_{in} = N$. We take $K_{H} = K_{W} = 3$ in this work unless otherwise stated. The output feature map $\mathbf{Z}^{(l)} \in \mathrm{R}^{C_{out} \times H^{\prime} \times W^{\prime}}$ of the $l^{th}$ layer is given by

\begin{equation}
    \mathbf{Z}^{(l)} = \mathbf{W}^{(l)} \ast \mathbf{X}^{(l-1)},
    \label{eq:convolution}
\end{equation}

\noindent where the symbol $\ast$ denotes the discrete convolution operator. It slides the kernel $\mathbf{W}^{(l)}$ across $\mathbf{X}^{(l-1)}$ at a certain stride (1 in this work), performs an entry-wise multiplication, and sums up the products into a scalar. While the dimensions of $\mathbf{X}^{(l-1)}$ and $\mathbf{Z}^{(l)}$ can be different in general, we make $H^{\prime} = H$ and $W^{\prime} = W$ by introducing $\textit{padding}$ around the element patch, which basically enlarges the element patch by certain layers of zeros (1 layer in this work due to choice of $K_{H}$ and $K_{W}$).

$\mathbf{Z}^{(l)}$ usually goes through the \textit{Batch Normalization} (BN) layer, which normalizes $\mathbf{Z}^{(l)}$ to have zero mean and unit variance, and thus puts the parameters on a similar scale to enhance the convergence of training. It is defined as 
\begin{equation}
    \mathbf{F}^{(l)} =  \text{BN}\left( \mathbf{Z}^{(l)} \right),
    \label{eq:convolution and BN}
\end{equation}

\noindent where the network BN contains learnable weights accounting for scaling and shifting. $\mathbf{F}^{(l)}$ is followed by the ReLU activation and then fed into the next convolutional layer.

This unit of Conv, BN and ReLU repeats several times (2 in this work) in a single residual block. Moreover, a residual connection is used right before the last activation to deal with the vanishing gradients in deep networks. When the input and output dimensions of a residual block differ (e.g., $C_{out} \neq C_{in}$), an extra layer of Conv and BN will be added to the residual connection to maintain compatible dimensions. Specifically, it introduces a kernel $\mathbf{W}_{b} \in \mathrm{R}^{C_{out} \times C_{in} \times 1 \times 1}$ to act on $\mathbf{X}^{(l-1)}$. CFE in this work consists of 6 residual blocks described above, whose output dimensions are 64, 128, 256, 256, 128, and 64, respectively. After residual blocks, a convolutional layer is used to aggregate all the extracted features into a single latent vector.

Subsequently, the latent vector is fed into FCCH to predict the element density. FCCH simply consists of several fully connected dense layers. In this work, it has 3 dense layers. Each of the first 2 layers has 100 neurons with a ReLU activation, whereas the number of neurons in the last layer corresponds to the output dimension. Note that activation functions are not needed in the last layer. 

Following DLTOP, eCNNTO also treats density prediction as a classification task rather than a regression problem; see Equations \ref{Eq:classification criterion1} and \ref{Eq:classification criterion2}. Correspondingly, the cross-entropy loss is used,
\begin{equation}\label{Eq:Cross Entropy}
    L = - \sum_{i=1}^{K} \rho_{i} \ln{\hat{\rho}_{i}},
\end{equation}

\noindent where $K$ ($K=3$ or $12$) is the total number of classes, and $\hat{\rho}_{i}$ denotes the predicted probability of the target element that falls into the class $\rho_{i}$. The specific choice of $K$ and its influence will be discussed in \autoref{Sec4}. All the neural nets are trained using Adam with a learning rate of $10^{-3}$ and a weight decay of $10^{-4}$. Each model is trained with 100 epochs.

\begin{rmk}

    Compared with DLTOP, eCNNTO replaces the Deep Relief Networks (DBNs) with a ResNet-based CNN to preserve spatial information and thus enhance structural connectivity. Alternative network architectures, such as Transformers and Vision Transformers, were also investigated. However, Transformer architectures lack an explicit inductive bias for local spatial correlations, whereas Vision Transformers require much larger datasets than CNN to fully exploit the modeling capacity \citep{dosovitskiyImageWorth16x162021}. As a result, these architectures show an inferior performance compared with the proposed method.
    
\end{rmk}

\subsection{Dataset Construction}\label{Sec32}

\begin{figure*}
\centering

    \begin{subfigure}{0.45\textwidth}
        \includegraphics[width=\linewidth]{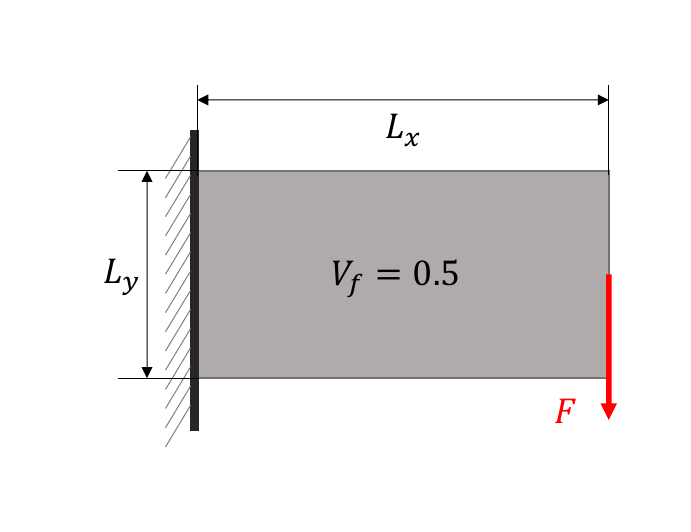}
        \caption{Cantilever beam}
        \label{fig:training benchmarks1}
    \end{subfigure}
    \hspace{0.2em}
    \begin{subfigure}{0.45\textwidth}
        \includegraphics[width=\linewidth]{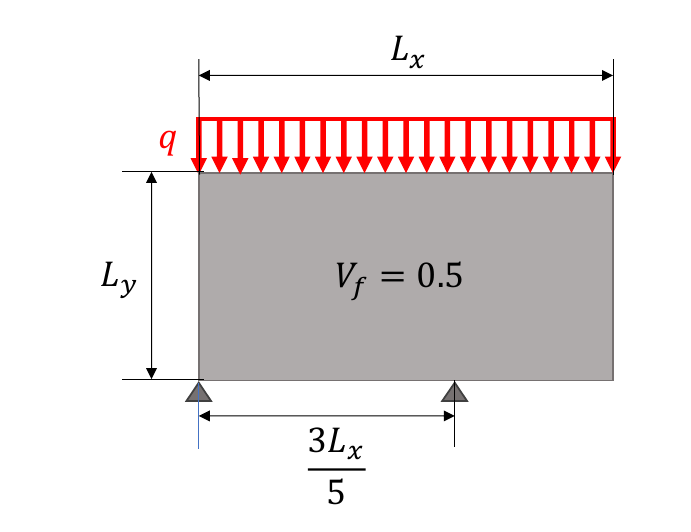}
        \caption{Simply supported beam}
        \label{fig:training benchmarks2}
    \end{subfigure}

\caption{Problem setups for dataset construction in 2D, where $L_{x}=2$, $L_{y} = 1$, and $V_{f}=0.5$ is the volume fraction. (a) A cantilever beam with a concentrated force $F=1$ applied at the midpoint of the right boundary, and (b) a simply supported beam with a uniformly distributed force $q=1$ applied on the top.}\label{fig:training benchmarks}
\end{figure*}

\begin{figure*}
\centering

    \begin{subfigure}{0.43\textwidth}
        \includegraphics[width=\linewidth]{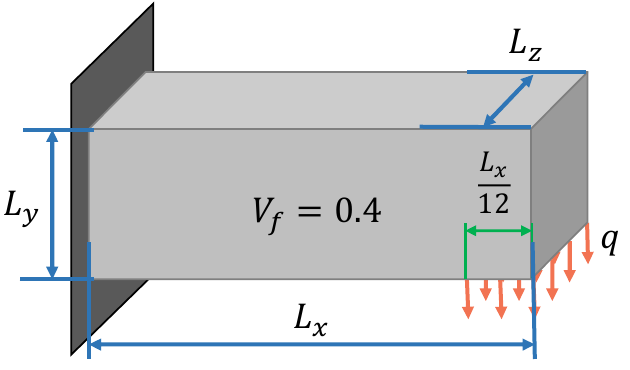}
        \caption{}
        \label{fig:3D training benchmarks1}
    \end{subfigure}
    \hspace{0.2em}
    \begin{subfigure}{0.47\textwidth}
        \includegraphics[width=\linewidth]{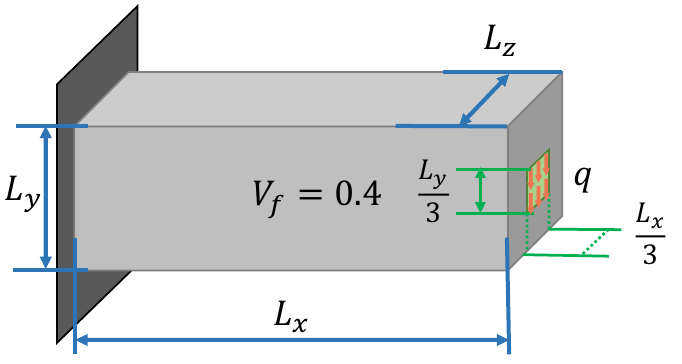}
        \caption{}
        \label{fig:3D training benchmarks2}
    \end{subfigure}
    
\caption{Problem setups for dataset construction in 3D, where $L_{x}=2$, $L_{y}=1$, $L_{z}=1$, $q=1$, and $V_{f}=0.4$ is the volume fraction. (a) A cantilever beam with a distributed force applied at a right area of the lower boundary, and (b) a cantilever beam with a distributed force applied at a center area of the right boundary.}\label{fig:3D training benchmarks}
\end{figure*}

Datasets are prepared by performing SIMP only on a couple of benchmark problems in 2D and 3D; see Figures \ref{fig:training benchmarks} and \ref{fig:3D training benchmarks}, where the problem setups are intended to be limited and simple in terms of geometries, meshes, and loading cases because later we would like to show the generalization capabilities of the proposed method. \autoref{fig:training benchmarks} shows the 2D case, which uses a cantilever beam and a simply supported beam. Two mesh resolutions are considered for a $2 \times 1$ design domain: $110 \times 60$ and $200 \times 100$. The 3D problem setups are shown in \autoref{fig:3D training benchmarks}. A $2 \times 1 \times 1$ cantilever beam is taken as the design domain with a single mesh resolution of $50 \times 100 \times 50$. Two loading cases are considered. 

Since the dataset is constructed at the element level, running a single problem can already yield a large number of data samples, so data generation is highly efficient. Indeed, through the above problems, we can obtain 53,200 samples in 2D and 500,000 in 3D, which are sufficient for training.

SIMP iterations are performed using open-source packages, including \cite{andreassenEfficientTopologyOptimization2011} for 2D problems and \cite{wangEfficientLargescale3D2025} for 3D. Note that we adopt the convergence criterion $\epsilon=1 \times 10^{-3}$ for 2D and $\epsilon=5 \times 10^{-3}$ for 3D, shown in \autoref{Eq:convergence criterion}. Parameters specific to SIMP are given in \autoref{tab:SIMP setting}.

\begin{table*}
\centering
\caption{Parameters setting of SIMP.}\label{tab:SIMP setting}
\begin{tabular}{cc} 
\toprule  
Parameter & Value\\
\midrule
Filter radius & 3\\
Penalty factor & 3\\
Young's modulus & 1\\
Minimum Young's modulus & $10^{-9}$\\
\bottomrule 
\end{tabular}
\end{table*}

\begin{rmk}
    Recall that the input is given in terms of an element patch. However, a boundary element does not have neighbors beyond the boundary. In this case, the element patch is padded with values of -1 to indicate that the corresponding neighbors do not exist.
\end{rmk}

\subsection{Feature selection for training}\label{Sec33}

\begin{figure*}
\centering
\includegraphics[width=0.9\textwidth]{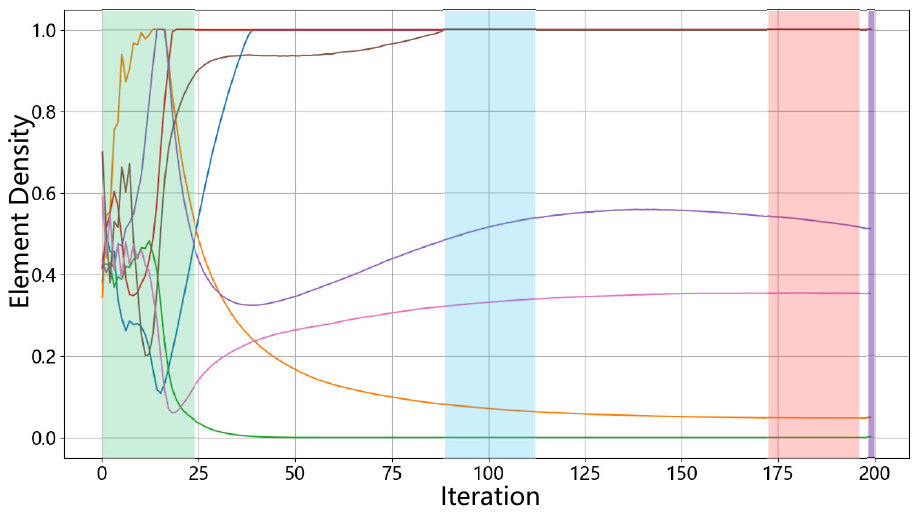}
\caption{Three options to select features for training: Early Stage (green), Middle Stage (blue), and Final Stage (red). A feature is a density sequence of consecutive SIMP iterations. Regardless of the feature choice, the label is always the final element density (purple).}\label{fig:training stage}
\end{figure*}

A data sample contains the entire density evolution of a given element patch obtained from SIMP iterations. For the training purpose, certain consecutive $N$ iterations are selected as the input feature, whereas the element density at the final iteration serves as the label. Depending on where these $N$ iterations come from, we investigate three options: the first $N$ iterations (Early Stage), the $N$ iterations in the middle of the evolution history (Middle Stage), and the last $N$ iterations except the final iteration (Final Stage); see \autoref{fig:training stage}. 

As will be shown, different choice of the feature leads to greatly varying performance in both training and testing. Note that DLTOP only considered the Early Stage strategy. Indeed, both the Middle and Final Stage strategies appear counter-intuitive. However, the Final Stage proves to over-perform the Early Stage in every aspect, such as the speedup for TO, structural connectivity, and the required data size.

To identify the most effective strategy, three neural nets with the identical network architecture and hyperparameters are trained on the same dataset (i.e., the 2D case introduced in \autoref{Sec32}). The only difference is the choice of the feature. Once trained, the neural nets are tested on three benchmark problems (see \ref{secA1} for details) in terms of the speedup and structural connectivity. \autoref{tab:iterations} lists the total number of iterations required to achieve an optimized structure. The results of ``SIMP'' serve as a reference, whereas the results of ``Early'', ``Middle'', and ``Final'' are obtained by applying the online prediction procedure of eCNNTO; see \autoref{fig:Overview framework}. ``NA'' means that the predicted structure is invalid due to the presence of disconnected components.

\begin{table*}
\centering
\caption{Total number of iterations required to achieve an optimized structure under different training strategies.}\label{tab:iterations}
\begin{tabular}{cccccc} 
\toprule  
\multirow{2}{*}{Example} & \multirow{2}{*}{Mesh} & \multicolumn{4}{c}{Number of iterations} \\
\cmidrule{3-6}        
& & SIMP & Early & Middle & Final \\
\midrule  
1 & \multirow{3}{*}{$80 \times 40$} & 429 & NA & NA & 155 \\
2 & & 112 & NA & 75 & 37 \\
3 & & 148 & 66 & 1024 & 63 \\
\bottomrule 
\end{tabular}
\end{table*}

We observe that the model trained with the Final Stage features consistently achieve the best acceleration performance and does not produce disconnected structures in any of the tested examples. In contrast, models trained with the other two options exhibit an inferior acceleration performance and may generate disconnected structures. Therefore, we will adopt the Final Stage features for training in this work.

It is worth mentioning that, once the model is trained and when it is used to accelerate TO, the input to the model is still given by the first $N$ SIMP iterations. Although it introduces a mismatch between the training and the inference of a model, the observed performance improvement can be explained from the characteristic of data. The data from Early Stage contains a large amount of noise due to the drastic evolution of the initial structure. The model trained on such data has difficulty in learning the density evolution pattern. In contrast, density variations in the Final Stage contain subtle yet critical information about the convergence behavior towards the optimized value, which greatly affects the acceleration performance. Accurately capturing this convergence behavior can help reduce the number of SIMP iterations required to fine-tune the near-optimal structure predicted by eCNNTO. On the other hand, the Middle Stage data only contains intermediate density variations that may be very similar in different problems and thus fail to distinguish them as independent features. As a result, its prediction may be poor given a new scenario, which also explains why the model trained on this strategy may be even worse than the case not using any acceleration.

\begin{rmk}

    DLTOP uses the Early Stage data as the feature for training, whereas eCNNTO uses data from the Final Stage. As the Early Stage data is much noisier than the Final Stage, DLTOP demands a larger dataset for training than eCNNTO, which will be shown in the next section.
    
\end{rmk}

\section{Numerical examples}\label{Sec4}

In this section, we present a variety of 2D and 3D test examples to demonstrate the efficiency and generalization capabilities of eCNNTO. The models have been trained according to the method introduced in the previous section. Here, they are directly applied to accelerate TO without retraining. We first investigate the effect of the classification criterion on both the acceleration performance and generalization of eCNNTO. Next, we compare eCNNTO with DLTOP, in particular on the structural connectivity and training data size. Last but not least, we evaluate the generalization capabilities of eCNNTO under various 2D and 3D settings, such as different boundary conditions, loading cases, design domain geometries, mesh resolutions, and non-design domains. 2D models are trained on an Intel\textsuperscript{\textregistered} Core\textsuperscript{\texttrademark} i7-12700 CPU @ 2.10 GHz with 64.0 GB DDR5 RAM and NVIDIA RTX A4000, whereas 3D models are trained on an Intel\textsuperscript{\textregistered} Xeon\textsuperscript{\textregistered} Gold 6248R CPU @ 3.00GHz with 32.0 GB DDR4 RAM and NVIDIA RTX A6000.

\subsection{Classification criterion}\label{Sec41}

\begin{figure*}
\centering

    \begin{subfigure}{0.45\textwidth}
        \includegraphics[width=\linewidth]{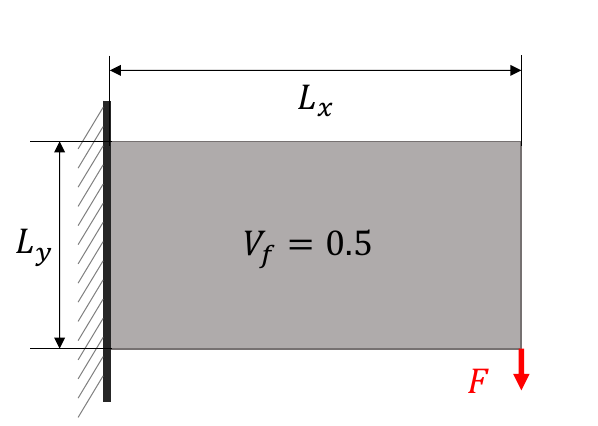}
        \caption{Cantilever beam}
        \label{fig:OOD1}
    \end{subfigure}
    \hspace{0.2em}
    \begin{subfigure}{0.45\textwidth}
        \includegraphics[width=\linewidth]{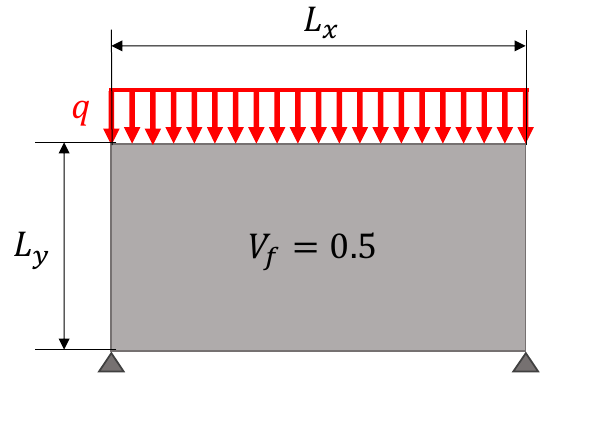}
        \caption{Simply supported beam}
        \label{fig:OOD2}
    \end{subfigure}

\caption{Two out-of-distribution examples, where $L_{x}=2$, $L_{y}=1$, and $V_{f}=0.5$ is the volume fraction. (a) A cantilever beam subjected to a concentrated force $F=1$ at its lower-right corner, and (b) a simply supported beam under a uniformly distributed force $q=1$.}\label{fig:OOD}
\end{figure*}

This section evaluates the impact of the classification granularity (i.e., the number of density classes) on the acceleration performance of eCNNTO. We compare two cases, where density values are classified into 3 classes or 12 classes; \autoref{Eq:classification criterion1} and \ref{Eq:classification criterion2}. Four examples are evaluated on a $2 \times 1$ design domain with a $110 \times 60$ mesh. Examples 1 and 2 are in-distribution (ID) samples drawn directly from the training dataset (from \autoref{fig:training benchmarks}), where the model is trained with the Final Stage sequence and here it is tested on the Early Stage sequence as the input. Examples 3 and 4 are out-of-distribution (OOD) problems featuring boundary conditions that are not covered during training; see \autoref{fig:OOD}. Both of them adopt the sensitivity filter with a radius of 3.

The results are summarized in \autoref{tab:effectclassification}. The number of iterations in eCNNTO (also DLTOP) is the sum of two parts: (1) the number of SIMP iterations used to generate the input, and (2) the number of SIMP iterations required to fine-tune the predicted structure by eCNNTO until convergence. We observe that both classification schemes yield substantial acceleration compared to the baseline SIMP across all the examples. Meanwhile, the 3-class scheme outperforms the 12-class one except for Example 1.

\begin{table*}
    \centering
    \caption{Comparison of two classification schemes in terms of the acceleration performance.}
    \label{tab:effectclassification}
    \begin{threeparttable}
    \begin{tabular}{lcccc}
        \toprule
        \multirow{2}{*}{Example} & \multicolumn{2}{c}{\# Iterations} & \multirow{2}{*}{\# Classes} & \multirow{2}{*}{Acceleration (\%)} \\
        \cmidrule(lr){2-3}
        & SIMP & eCNNTO & & \\
        \midrule
        \multirow{2}{*}{Example 1 (ID)} & \multirow{2}{*}{469} & 81 & 3 & 82.7 \\
        & & 62 & 12 & 86.8 \\
        \multirow{2}{*}{Example 2 (ID)} & \multirow{2}{*}{617} & 75 & 3 & 87.8 \\
        & & 103 & 12 & 83.3 \\
        \multirow{2}{*}{Example 3 (OOD)} & \multirow{2}{*}{494} & 88 & 3 & 82.2 \\
        & & 144 & 12 & 70.9 \\
        \multirow{2}{*}{Example 4 (OOD)} & \multirow{2}{*}{855} & 142 & 3 & 83.4 \\
        & & 186 & 12 & 78.2 \\
        \bottomrule
    \end{tabular}
    \begin{tablenotes}
    \footnotesize
    \item Note: Acceleration = (\# SIMP - \# eCNNTO) / \# SIMP $\times 100\%$ 
    \end{tablenotes}
\end{threeparttable}
\end{table*}

Intuitively, a finer classification granularity (e.g., 12 classes) narrows the density intervals, allowing the network predictions to more closely approximate the ground-truth labels and thus to accelerate convergence. However, a larger number of classes inherently increases the complexity of the classification task and makes it prone to falling into adjacent classes rather the target, leading to increased prediction errors on unseen data. This is the reason for the degraded performance of the 12-class scheme on OOD samples. On the other hand, the 3-class scheme shows a better trade-off between robustness and generalization capabilities across different problems, and is thus adopted for all the subsequent examples.

\subsection{Comparison with DLTOP}\label{Sec42}

\begin{figure*}
\centering

    \begin{subfigure}{0.3\textwidth}
        \includegraphics[width=\linewidth]{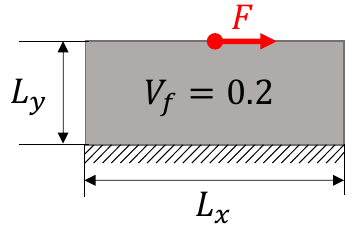}
        \caption{Problem setting}
        \label{fig:DLTOPCase11}
    \end{subfigure}
    \hfill
    \begin{subfigure}{0.33\textwidth}
        \includegraphics[width=\linewidth]{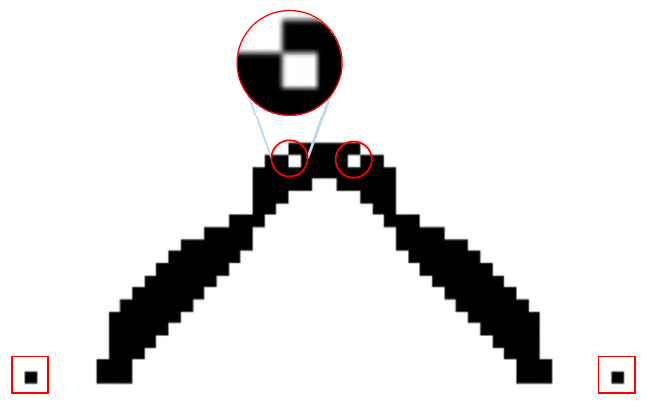}
        \captionsetup{skip=17pt}
        \caption{DLTOP}
        \label{fig:DLTOPCase12}
    \end{subfigure}
    \hfill
    \begin{subfigure}{0.33\textwidth}
        \includegraphics[width=\linewidth]{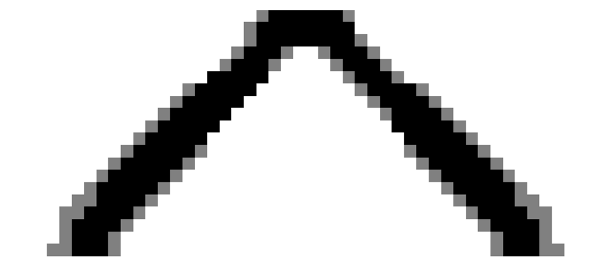}
        \captionsetup{skip=17pt}
        \caption{eCNNTO}
        \label{fig:DLTOPCase13}
    \end{subfigure}

\caption{Comparison of predicted structures between DLTOP and eCNNTO, where the optimized result should have a two-bar structure.  (a) A beam fixed at the bottom with a concentrated force $F=1$ loaded horizontally on the top, where $L_{x}=50$, $L_{y}=20$ and $V_{f}=0.2$ is the volume fraction, (b) predicted structure of DLTOP that has both isolated pieces (bottom-left and bottom-right) and corner-contact parts (top), and (c) predicted structure of eCNNTO, where the desired structural connectivity is well preserved.}\label{fig:DLTOPCase1}
\end{figure*}

This section demonstrates the advantages of eCNNTO in ensuring the structural connectivity and validity. We compare its performance with DLTOP on two representative examples in 2D. The first example is shown in \autoref{fig:DLTOPCase1}(a), where a $50 \times 20$ mesh is used for a $50 \times 20$ domain. We set the volume fraction and the filter radius as 0.2 and 1.5, respectively, which follows a typical setting when using SIMP for this example. The number of input iterations is 5 for both DLTOP and eCNNTO as it is sufficient to capture the early drastic change in the density evolution. The predicted structures of DLTOP and eCNNTO are shown in \autoref{fig:DLTOPCase1}(b, c), respectively. The predicted structure of DLTOP has isolated pieces in the bottom-left and bottom-right regions (highlighted in red boxes). It also has corner-contact parts in the top region (highlighted in the red circle). This is due to the fact that DLTOP relies solely on individual elements for prediction, making its outputs vulnerable to early evolutionary fluctuations. In contrast, the spatial correlation of the neighboring structural evolution is built in eCNNTO through CNN. As a result, even if the density evolution of an element fluctuates drastically, the variation as a group of neighboring elements can be smoothed out and thus yield much more accurate predictions.

\begin{figure*}
\centering

    \begin{subfigure}{0.33\textwidth}
        \includegraphics[width=\linewidth]{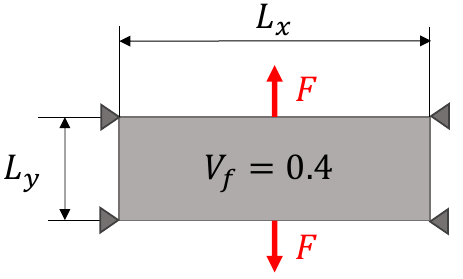}
        \captionsetup{skip=22pt}
        \caption{Problem setting}
        \label{fig:DLTOPCase21}
    \end{subfigure}
    \hfill
    \begin{subfigure}{0.30\textwidth}
        \includegraphics[width=\linewidth]{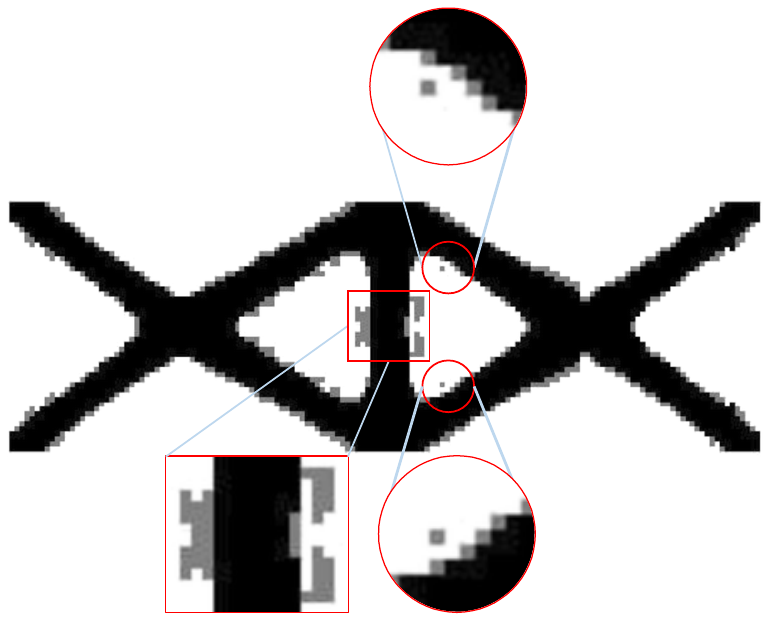}
        \caption{DLTOP}
        \label{fig:DLTOPCase22}
    \end{subfigure}
    \hfill
    \begin{subfigure}{0.30\textwidth}
        \includegraphics[width=\linewidth]{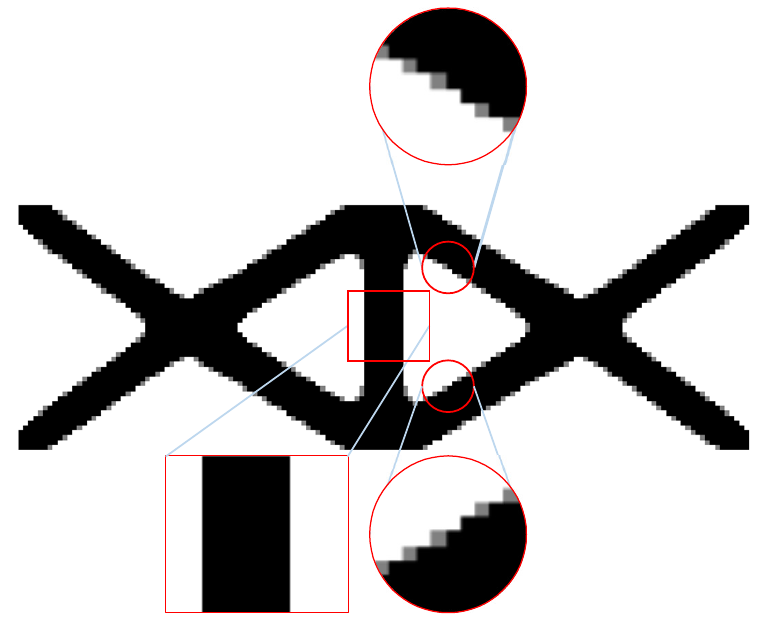}
        \caption{eCNNTO}
        \label{fig:DLTOPCase23}
    \end{subfigure}

\caption{Comparison of predicted structures between DLTOP and eCNNTO. (a) A beam simply supported at four corners with two concentrated forces $F=1$ loaded vertically on the top and bottom, where $L_{x}=3$, $L_{y}=1$, and $V_{f}=0.4$ is the volume fraction, (b) predicted structure of DLTOP that has isolated pieces (top and bottom red circles) and intermediate-density defects (middle red box), and (c) predicted structure of eCNNTO, where the desired structure is obtained without any defects.}\label{fig:DLTOPCase2}
\end{figure*}

Next, we study a more complex example shown in \autoref{fig:DLTOPCase2}(a), where we adopt a $150 \times 50$ mesh for a $3 \times 1$ domain, and set the volume fraction as 0.4. The sensitivity and density filters are used in SIMP with a radius of 2. The number of input iterations of both eCNNTO and DLTOP is 36. Similar results to the first example are observed in \autoref{fig:DLTOPCase2}(b, c), where isolated pieces (highlighted in the red circles) appear in the predicted structure of DLTOP but not in eCNNTO. Meanwhile, intermediate-density defects (gray shades in the red box) appear around the vertical component, where void is expected. This is because elements located at structural boundaries undergo drastic variations in densities, and the absence of spatial correlation leads to fluctuations in predicted densities near boundaries. In contrast, eCNNTO extracts density variation patterns of neighboring elements via CNN, and thus smooths out sharp fluctuations near structural boundaries. Therefore, it produces more accurate predictions for elements around boundaries.

Last but not least, eCNNTO achieves superior performance with much fewer training samples compared to DLTOP. It only requires 11.1\% of training samples needed in DLTOP: 53,200 (eCNNTO) versus 480,000 (DLTOP). Due to the lack of spatial correlations among elements, DLTOP demands higher training costs yet fails to effectively resolve isolated pieces and intermediate-density defects. Moreover, further enlarging the DLTOP dataset would not overcome this issue. In contrast, by adjusting the network architecture and carefully choosing features for training, eCNNTO generates structures with better connectivity and much fewer defects while significantly reducing the required data size. Moreover, these adjustments can further enhance the speedup of eCNNTO, which will be discussed in detail in \autoref{Sec43}.

\subsection{eCNNTO in 2D}\label{Sec43}

\begin{figure*} 
    \centering
    \begin{subfigure}{0.38\textwidth}
        \includegraphics[width=\linewidth]{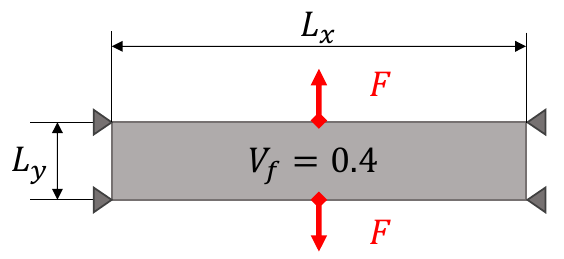}
        \caption{Long beam}
        \label{fig:2Dcase1}
    \end{subfigure}
    \hfill
    \begin{subfigure}{0.26\textwidth}
        \includegraphics[width=\linewidth]{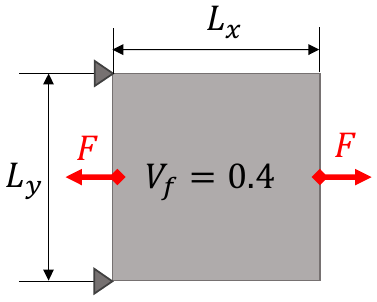}
        \caption{Square}
        \label{fig:2Dcase2}
    \end{subfigure}
    \hfill
    \begin{subfigure}{0.23\textwidth}
        \includegraphics[width=\linewidth]{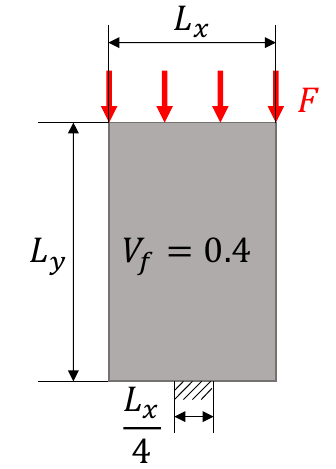}
        \caption{Column}
        \label{fig:2Dcase3}
    \end{subfigure}

    \vspace{1em}

    \begin{subfigure}{0.35\textwidth}
        \includegraphics[width=\linewidth]{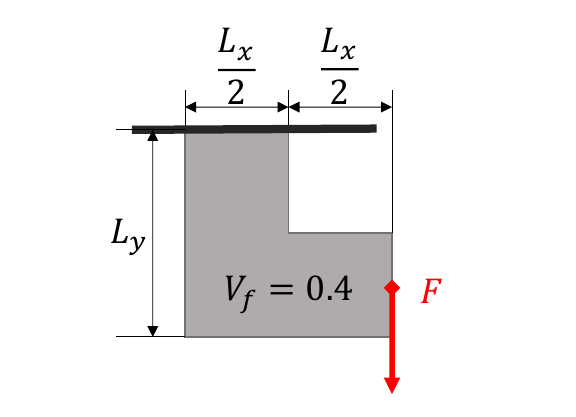}
        \caption{L-shaped}
        \label{fig:2Dcase4}
    \end{subfigure}
    \hspace{0.2em}
    \begin{subfigure}{0.35\textwidth}
        \includegraphics[width=\linewidth]{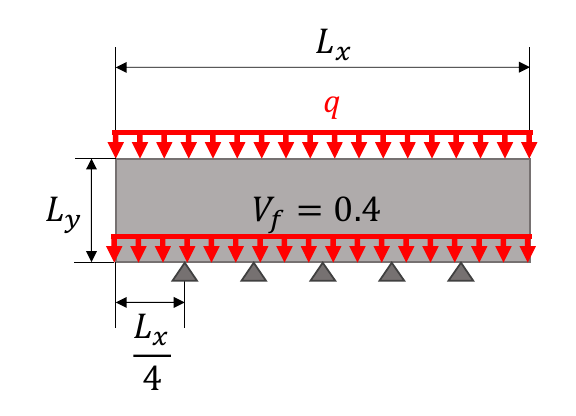}
        \caption{Uniformly distributed loading (UDL)}
        \label{fig:2Dcase5}
    \end{subfigure}
    \caption{Problem settings of 2D test problems, where $V_{f}=0.4$ is the volume fraction. (a) A long beam simply supported at four corners with $L_{x}=8$, $L_{y}=1.5$, and $F=1$, (b) a square domain simply supported at the left two corners with $L_{x}=4$, $L_{y}=4$, and $F=1$, (c) a vertical column partially fixed at the bottom with $L_{x}=3$, $L_{y}=5$, and $F=1$, (d) an L-shaped bracket clamped on the top with $L_{x}=2$, $L_{y}=2$, and $F=1$, and (e) a beam simply supported at five places on the bottom and applied with two uniformly distributed forces $q=1$, with $L_{x}=4$ and $L_{y}=1$.}
    \label{fig:2Dcase}
\end{figure*}

This section evaluates the efficiency and the generalizability of eCNNTO using 2D problems. The model has been trained according to the problem settings in \autoref{fig:training benchmarks}. Now it is tested directly on the five problems shown in \autoref{fig:2Dcase} without any kind of retraining.  Comparing \autoref{fig:training benchmarks} and \ref{fig:2Dcase}, we observe the significant difference between the training data and the test cases in terms of boundary conditions, loading cases, and design domain geometries. In fact, different mesh resolutions will also be tested; see \autoref{table:2D mesh}. These tests are mainly intended to show the generalization capabilities of eCNNTO. Regarding TO parameters, the volume fraction of all these examples are $0.4$. The radius of sensitivity filter is 6 for all but the L-shaped beam, which takes~2. For eCNNTO, the window size of an element patch is $W=5$ and the number of input iterations is $N=48$.

\begin{table*}
\caption{Mesh resolutions of 2D test problems.}\label{table:2D mesh}
\centering
\begin{threeparttable}
\setlength{\tabcolsep}{40pt}
\begin{tabular}{lcc}
\toprule
Example & $N_{x}$  & $N_{y}$\\
\midrule
Long beam  & 800   & 150 \\
Square     & 400   & 400 \\
Column     & 300   & 500 \\
L-shaped   & 400   & 400 \\
UDL   & 400   & 100 \\
\bottomrule
\end{tabular}
\vspace{4pt}
\begin{tablenotes}
    \footnotesize
    \item Note: $N_{x}$ and $N_{y}$ are the number of elements in the x and y directions, respectively. In the case of the L-shaped domain, the mesh is meant for its bounding box.
\end{tablenotes}
\end{threeparttable}
\end{table*}

\begin{table*}
\caption{Acceleration performance of eCNNTO in 2D and resulting compliance.}\label{table:2D result}%
\centering
\setlength{\tabcolsep}{3pt}
\begin{tabular}{@{}lccccc@{}}
\toprule
\multirow{2}{*}{Example} & \multicolumn{2}{c}{\# Iterations} & \multirow{2}{*}{Acceleration (\%)} & \multicolumn{2}{c}{Compliance}\\
\cmidrule(lr){2-3} \cmidrule(lr){5-6}
& SIMP & eCNNTO & & SIMP & eCNNTO\\
\midrule
Long beam     & 469  & 71  & 84.9 &86.32  & \textbf{80.46}\\
Square        & 935  & 226 & 75.8 &22.65  & \textbf{22.09}\\
Column        & 890  & 85  & 90.5 &147.44 & \textbf{143.64}\\
L-shaped      & 1009 & 205 & 79.7 &89.41  & 90.09\\
UDL           & 703  & 113 & 83.9 &1.33   & \textbf{1.18}\\
\bottomrule
\end{tabular}
\end{table*}

\begin{table*}
\centering
\caption{Runtime comparison in 2D test problems between SIMP and eCNNTO.}\label{table:2D time comparison}%
\begin{tabular}{@{}lccc@{}}
\toprule
\multirow{2}{*}{Example} & \multicolumn{2}{c}{Time (s)} & \multirow{2}{*}{Acceleration (\%)}\\ 
\cmidrule(lr){2-3}
& SIMP & eCNNTO \\
\midrule
Long beam   & 4773.7  & 716.7  & \textbf{85.0}\\
Square      & 21630.1 & 5193.7 & \textbf{76.0}\\
Column      & 64078.9 & 5308.1 & \textbf{91.7}\\
L-shaped    & 6425.4  & 1273.6 & \textbf{80.2}\\
UDL    & 1319.5  & 173.9  & 86.8\\
\bottomrule
\end{tabular}
\end{table*}

\begin{figure*}
    \centering

    \begin{minipage}[c]{0.42\textwidth}
        \centering
        \begin{subfigure}{\linewidth}
            \centering
            \includegraphics[width=0.95\linewidth]{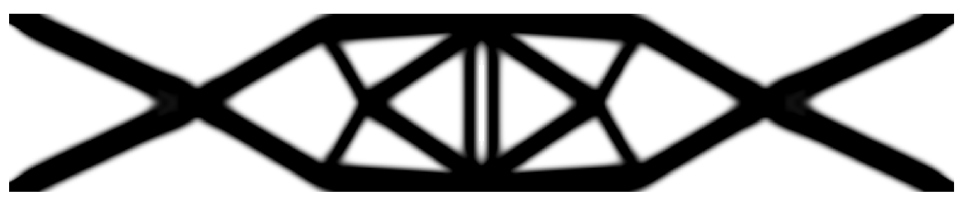}
            \caption{SIMP}
            \label{fig:2Dstructurescase11}
        \end{subfigure}
        
        \vspace{1em}
        
        \begin{subfigure}{\linewidth}
            \centering
            \includegraphics[width=0.95\linewidth]{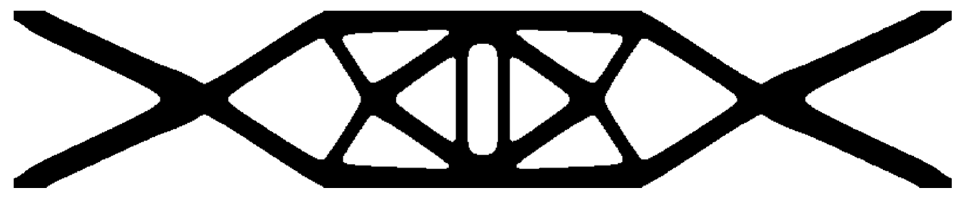}
            \caption{eCNNTO}
            \label{fig:2Dstructurescase12}
        \end{subfigure}
    \end{minipage}
    \hfill
    \begin{minipage}[c]{0.48\textwidth}
        \centering
        \begin{subfigure}{0.48\linewidth}
            \centering
            \includegraphics[width=\linewidth]{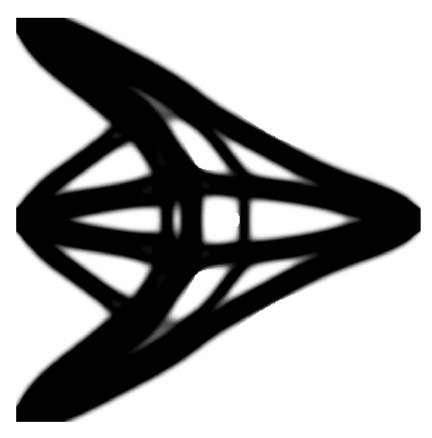}
            \caption{SIMP}
            \label{fig:2Dstructurescase21}
        \end{subfigure}
        \hfill
        \begin{subfigure}{0.48\linewidth}
            \centering
            \includegraphics[width=\linewidth]{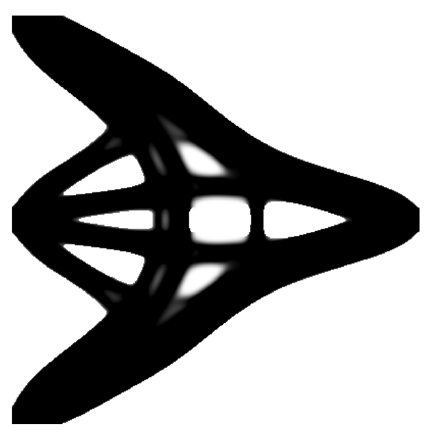}
            \caption{eCNNTO}
            \label{fig:2Dstructurescase22}
        \end{subfigure}
    \end{minipage}

    \vspace{2.5em}

    \begin{minipage}[c]{0.33\textwidth}
        \centering
        \begin{subfigure}{0.48\linewidth}
            \centering
            \includegraphics[width=\linewidth]{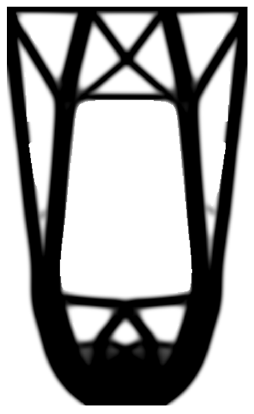}
            \caption{SIMP}
            \label{fig:2Dstructurescase31}
        \end{subfigure}
        \hfill
        \begin{subfigure}{0.48\linewidth}
            \centering
            \includegraphics[width=\linewidth]{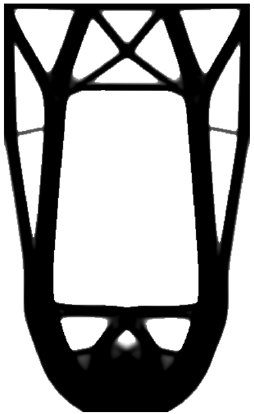}
            \caption{eCNNTO}
            \label{fig:2Dstructurescase32}
        \end{subfigure}
    \end{minipage}
    \hfill
    \begin{minipage}[c]{0.49\textwidth}
        \centering
        \begin{subfigure}{0.48\linewidth}
            \centering
            \includegraphics[width=\linewidth]{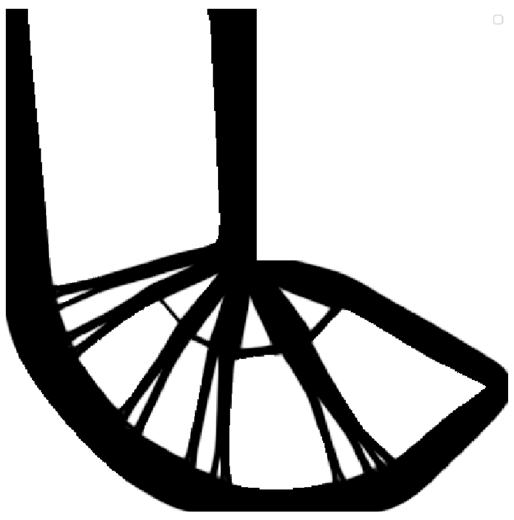}
            \caption{SIMP}
            \label{fig:2Dstructurescase41}
        \end{subfigure}
        \hfill
        \begin{subfigure}{0.48\linewidth}
            \centering
            \includegraphics[width=\linewidth]{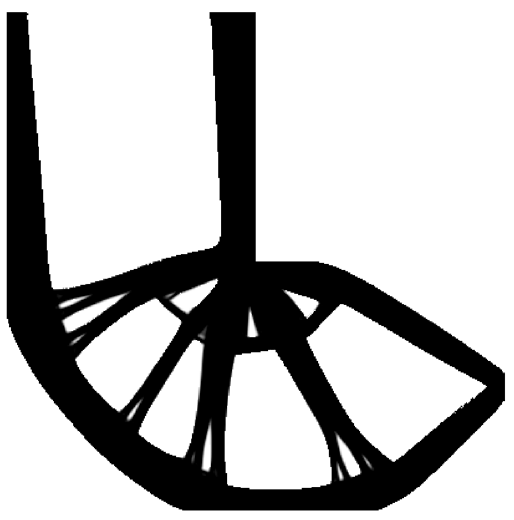}
            \caption{eCNNTO}
            \label{fig:2Dstructurescase42}
        \end{subfigure}
    \end{minipage}

    \vspace{2.5em}

    \begin{minipage}[c]{0.60\textwidth}
        \centering
        \begin{subfigure}{\linewidth}
            \centering
            \includegraphics[width=\linewidth]{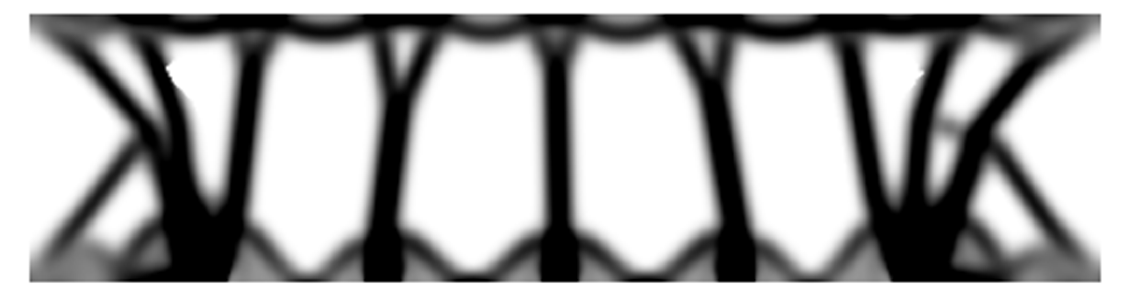}
            \caption{SIMP}
            \label{fig:2Dstructurescase51}
        \end{subfigure}
        
        \vspace{1em}
        
        \begin{subfigure}{\linewidth}
            \centering
            \includegraphics[width=\linewidth]{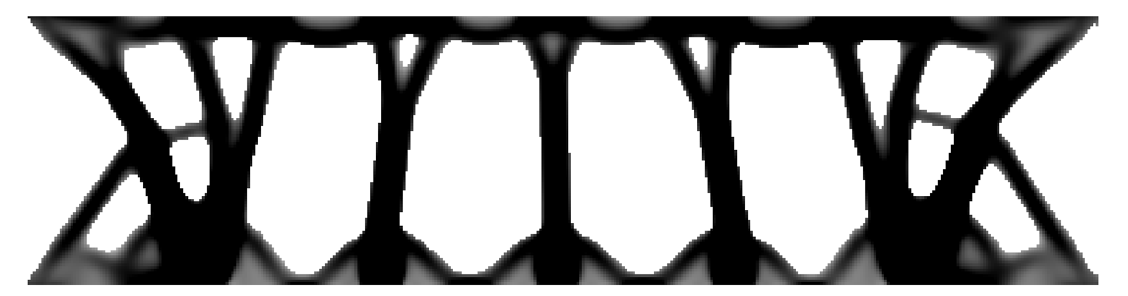}
            \caption{eCNNTO}
            \label{fig:2Dstructurescase52}
        \end{subfigure}
    \end{minipage}

    \caption{Optimized structures in 2D using SIMP and eCNNTO: (a, b) Long beam, (c, d) Square, (e, f) Column, (g, h) L-shaped, and (i, j) UDL. In each pair, the former is obtained by SIMP and the latter by eCNNTO.}
    \label{fig:alloptimizedstructures}
\end{figure*}

The acceleration performance and the corresponding structural compliance (i.e., the optimization objective) are summarized in \autoref{table:2D result}, where the SIMP results are taken as a baseline. We first observe that eCNNTO achieves a substantial reduction in the number of SIMP iterations across all the five examples, with the acceleration ratio ranging from 75.8\% to 90.5\%. This superior performance demonstrates the strong generalization capability of eCNNTO:  without retraining, eCNNTO yields significant speedup under unseen boundary conditions, loading cases, domain geometries, and mesh resolutions. Moreover, eCNNTO produces compliance comparable to, and in most examples lower than, those obtained by SIMP, demonstrating that it can significantly accelerate convergence without compromising the performance. This is because the predicted structure by eCNNTO is fine-tuned by SIMP to meet the same convergence criterion, which yields the compliance comparable to that of SIMP. 

The optimized structures are shown in \autoref{fig:alloptimizedstructures}, where the structures of eCNNTO exhibit clearer boundaries and fewer intermediate-density regions. While the optimized structures of SIMP and eCNNTO show discrepancies in certain structural details, the compliance values of eCNNTO are usually smaller. This is due to the fact that topology optimization is non-convex. The predicted structure by eCNNTO generally does not correspond to a certain iteration of the original SIMP method, thereby leading to slightly different but close local minima. We find that increasing the number of input iterations can effectively mitigate this difference.

Regarding the runtime, it almost scales linearly with respect to the number of SIMP iterations, so the reduction of SIMP iterations by eCNNTO leads to significant speedup in the actual computational time; see \autoref{table:2D time comparison}. We observe that the acceleration ratio in runtime ranges from 76.0\% to 91.7\%, which shows an order-of-magnitude speedup.

We further compare eCNNTO with DLTOP in terms of the acceleration performance. The acceleration ratios (computed against SIMP) of both methods are reported in \autoref{table:2D comparison}. We observe that eCNNTO consistently outperforms DLTOP in all examples. Note that the results of DLTOP come from \cite{kalliorasAcceleratedTopologyOptimization2020}.

\begin{table*}
\centering
\setlength{\tabcolsep}{15pt}
\caption{2D optimization comparison between DLTOP and eCNNTO.}\label{table:2D comparison}%
\begin{tabular}{lcc}
\toprule
\multirow{2}{*}{Example} & \multicolumn{2}{c}{Acceleration (\%)}\\
\cmidrule(lr){2-3}
& DLTOP & eCNNTO \\
\midrule
Long beam    & 73.5 & \textbf{84.9}\\
Square       & 56.7 & \textbf{75.8}\\
Column       & 84.3 & \textbf{90.5}\\
L-shaped     & 72.5 & \textbf{79.7}\\
UDL          & 66.0 & \textbf{83.9}\\
\bottomrule
\end{tabular}
\end{table*}

\subsection{eCNNTO in 3D}\label{Sec44}

\begin{figure*}
\centering

    \begin{subfigure}{0.48\textwidth}
        \includegraphics[width=\linewidth]{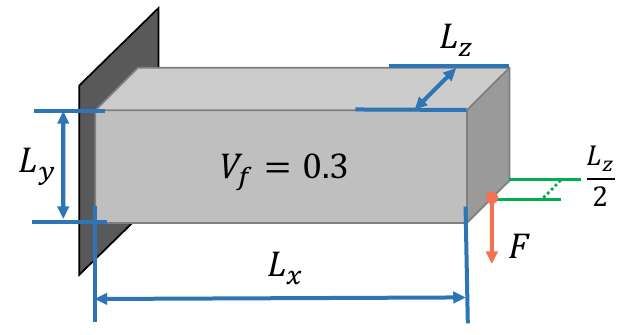}
        \caption{Cantilever 1}
        \label{fig:3Dcase1}
    \end{subfigure}
    \hfill 
    \begin{subfigure}{0.48\textwidth}
        \includegraphics[width=\linewidth]{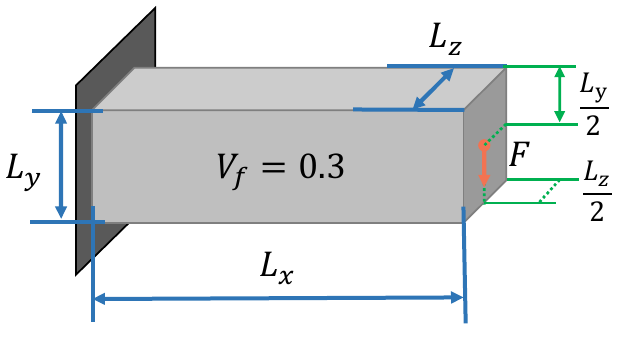}
        \caption{Cantilever 2}
        \label{fig:3Dcase2}
    \end{subfigure}

    \vspace{1em}

    \begin{subfigure}{0.58\textwidth}
        \includegraphics[width=\linewidth]{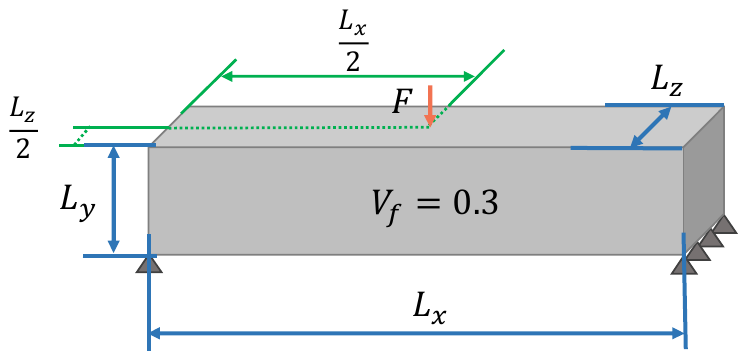}
        \caption{Long beam}
        \label{fig:3Dcase3}
    \end{subfigure}
    \hfill 
    \begin{subfigure}{0.38\textwidth}
        \includegraphics[width=\linewidth]{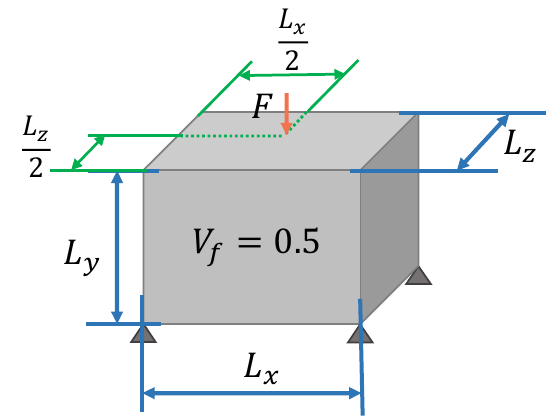}
        \caption{Short beam}
        \label{fig:3Dcase4}
    \end{subfigure}
    
\caption{Problem settings of 3D test problems, where $F=1$ is the concentrated force. (a) A cantilever with a concentrated force at the bottom of the right boundary with $L_{x}=200$, $L_y=100$, and $L_z=100$, (b) a cantilever with a concentrated force at the center of the right boundary with $L_{x}=200$, $L_y=100$, and $L_z=100$, (c) a simply supported long beam with a concentrated force at the center of the top surface with $L_{x}=300$, $L_y=100$, and $L_z=100$, and (d) a simply supported short beam with a concentrated force at the center of the top surface with $L_{x}=200$, $L_y=200$, and $L_z=200$.}\label{fig:3Dcase}
\end{figure*}

This section evaluates eCNNTO on 3D test problems to demonstrate its significant speedup. The model has been trained on the problem setttings in \autoref{fig:3D training benchmarks}, whereas four problems shown in \autoref{fig:3Dcase} are tested without retraining. Similar to the 2D scenario, these test problems have different boundary conditions and loading cases from the training examples, as can be seen by comparing Figures \ref{fig:3D training benchmarks} and \ref{fig:3Dcase}. Moreover, the mesh resolutions of the test problems are significantly larger than those of the training problems, as shown in \autoref{table:3D mesh}. Recall that a $50 \times 100 \times 50$ mesh is used for training. For all 3D examples, a window size of $W=3$, the number of iterations $N=24$, and a PDE filter \citep{kawamotoHeavisideProjectionBased2011} with radius 3 are adopted. The other hyperparameters and training settings are kept consistent with those in the 2D experiments.

\begin{table*}
\caption{Mesh resolutions of 3D test problems.}\label{table:3D mesh}
\centering
\begin{threeparttable}
\setlength{\tabcolsep}{20pt}
\begin{tabular}{lccc}
\toprule
Example & $N_{x}$  & $N_{y}$ & $N_{z}$\\
\midrule
Cantilever 1  & 200   & 100 & 100 \\
Cantilever 2  & 200   & 100 & 100 \\
Long beam     & 300   & 100 & 100 \\
Short beam    & 200   & 200 & 200 \\
\bottomrule
\end{tabular}
\vspace{4pt}
\begin{tablenotes}
    \footnotesize
    \item Note: $N_{x}$, $N_{y}$ and $N_{z}$ are the number of elements in the x, y, and z directions, respectively.
\end{tablenotes}
\end{threeparttable}
\end{table*}

\begin{table*}
\caption{Acceleration performance of eCNNTO in 3D and resulting compliance}\label{table:3D result}%
\centering
\setlength{\tabcolsep}{3pt}
\begin{tabular}{@{}lcccccc@{}}
\toprule
\multirow{2}{*}{Example} & \multicolumn{2}{c}{\# Iterations} & \multirow{2}{*}{Acceleration (\%)} & \multicolumn{2}{c}{Compliance}\\
\cmidrule(lr){2-3} \cmidrule(lr){5-6}
& SIMP & eCNNTO & & SIMP & eCNNTO\\
\midrule
Cantilever 1 & 376         & 83 & 77.9        &4.70    & \textbf{4.66}\\
Cantilever 2 & 317         & 76 & 76.0         &2.42   & \textbf{2.40}\\
Long beam    & $\geq 2000$ & 90  & $\geq 95.5$ &1.71   & \textbf{1.70}\\
Short beam   & $\geq 2000$ & 65  & $\geq 96.8$ &3.17   & 3.17 \\
\bottomrule
\end{tabular}
\end{table*}

\begin{table*}
\centering
\caption{Runtime comparison in 3D test problems between SIMP and eCNNTO.}\label{table:3D time comparison}%
\begin{tabular}{@{}lccc@{}}
\toprule
\multirow{2}{*}{Example} & \multicolumn{2}{c}{Time (s)} & \multirow{2}{*}{Acceleration (\%)}\\ 
\cmidrule(lr){2-3}
& SIMP & eCNNTO \\
\midrule
Cantilever 1    & 8805.9    & 1790.4  & 79.7\\
Cantilever 2    & 7317.6    & 1598.4  & 78.2\\
Long beam       & 50686.0   & 2548.2  & 95.0\\
Short beam      & 101346.0  & 5227.8  & 94.8\\
\bottomrule
\end{tabular}
\end{table*}

\begin{figure*}[htbp]
    \centering

    \begin{subfigure}{0.33\textwidth}
        \centering
        \includegraphics[width=\linewidth]{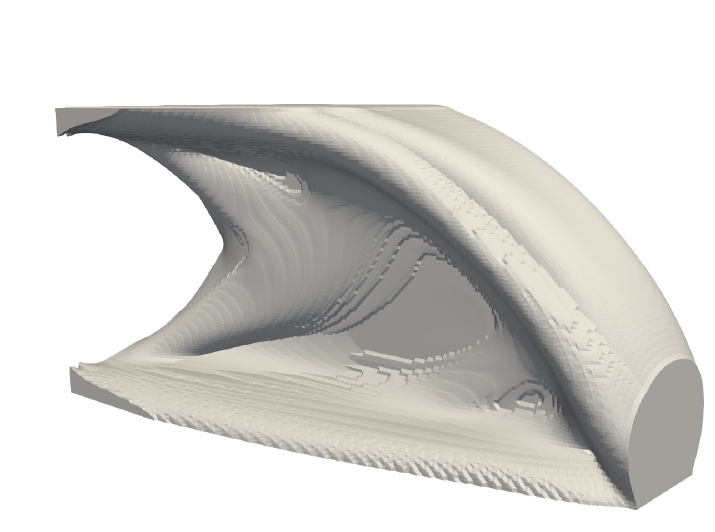}
        \caption{SIMP}
        \label{fig:3Dstructurescase11}
    \end{subfigure}
    \hspace{0.05\textwidth}
    \begin{subfigure}{0.33\textwidth}
        \centering
        \includegraphics[width=\linewidth]{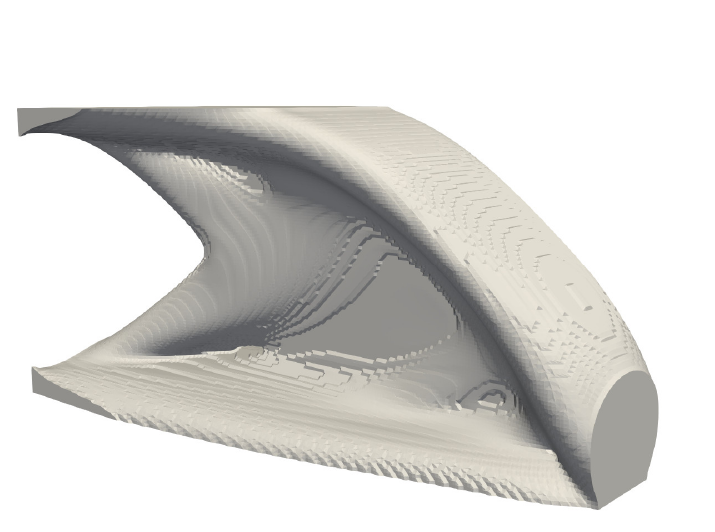}
        \caption{eCNNTO}
        \label{fig:3Dstructurescase12}
    \end{subfigure}

    \vspace{0.5em}

    \begin{subfigure}{0.33\textwidth}
        \centering
        \includegraphics[width=\linewidth]{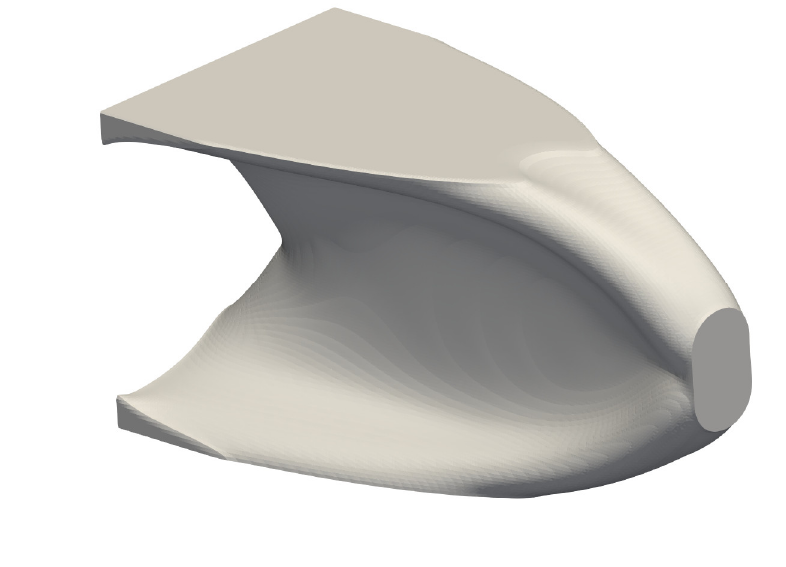}
        \caption{SIMP}
        \label{fig:3Dstructurescase21}
    \end{subfigure}
    \hspace{0.05\textwidth}
    \begin{subfigure}{0.33\textwidth}
        \centering
        \includegraphics[width=\linewidth]{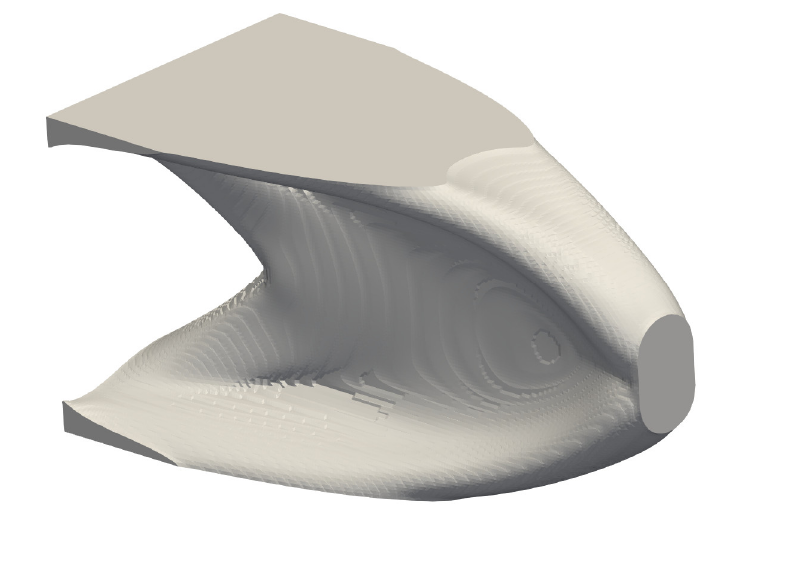}
        \caption{eCNNTO}
        \label{fig:3Dstructurescase22}
    \end{subfigure}

    \vspace{0.5em}

    \begin{subfigure}{0.33\textwidth}
        \centering
        \includegraphics[width=\linewidth]{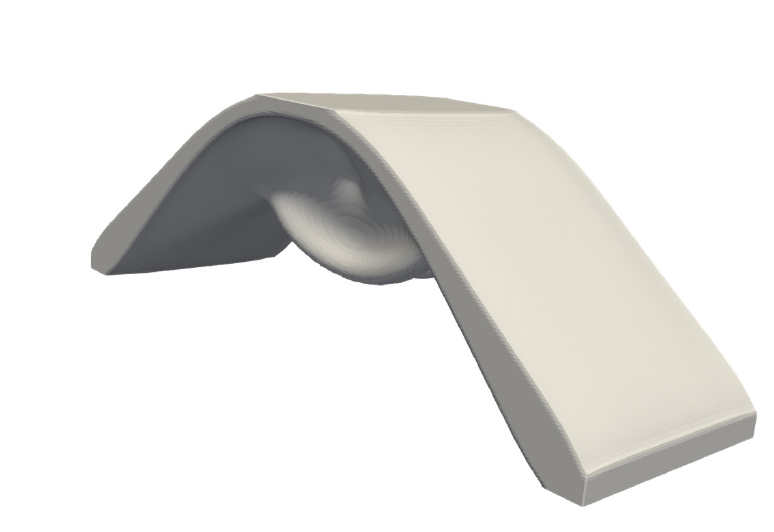}
        \caption{SIMP}
        \label{fig:3Dstructurescase31}
    \end{subfigure}
    \hspace{0.05\textwidth}
    \begin{subfigure}{0.33\textwidth}
        \centering
        \includegraphics[width=\linewidth]{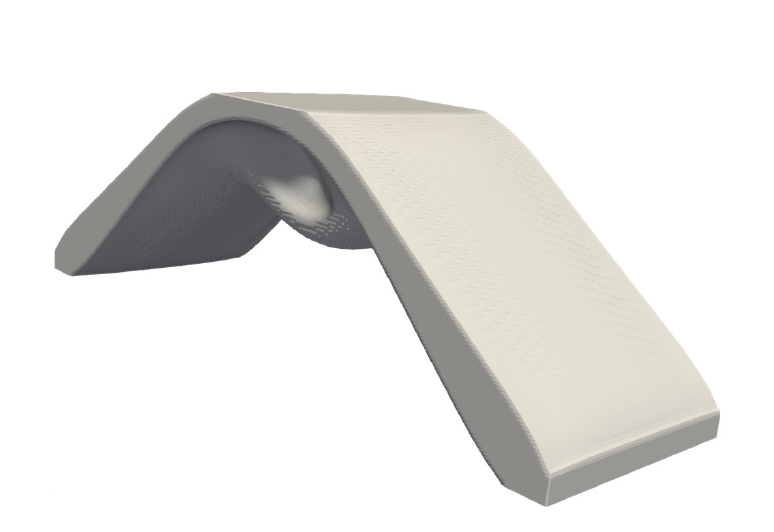}
        \caption{eCNNTO}
        \label{fig:3Dstructurescase32}
    \end{subfigure}

    \vspace{0.5em}
    
    \begin{subfigure}{0.33\textwidth}
        \centering
        \includegraphics[width=\linewidth]{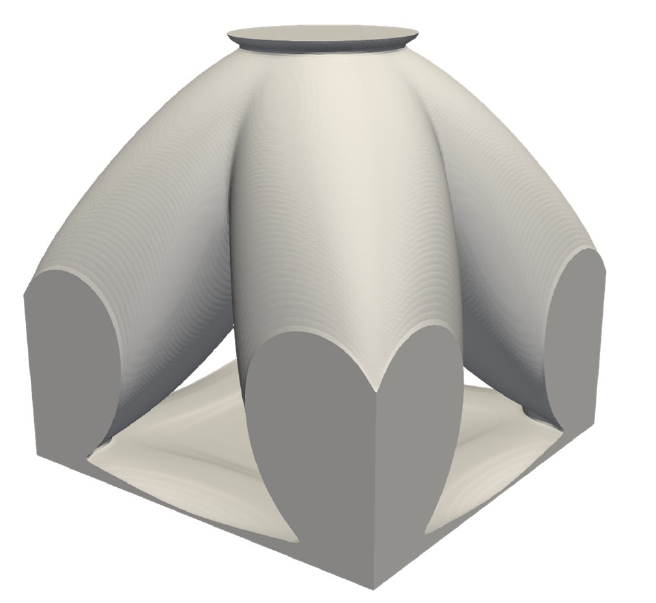}
        \caption{SIMP}
        \label{fig:3Dstructurescase41}
    \end{subfigure}
    \hspace{0.05\textwidth}
    \begin{subfigure}{0.33\textwidth}
        \centering
        \includegraphics[width=\linewidth]{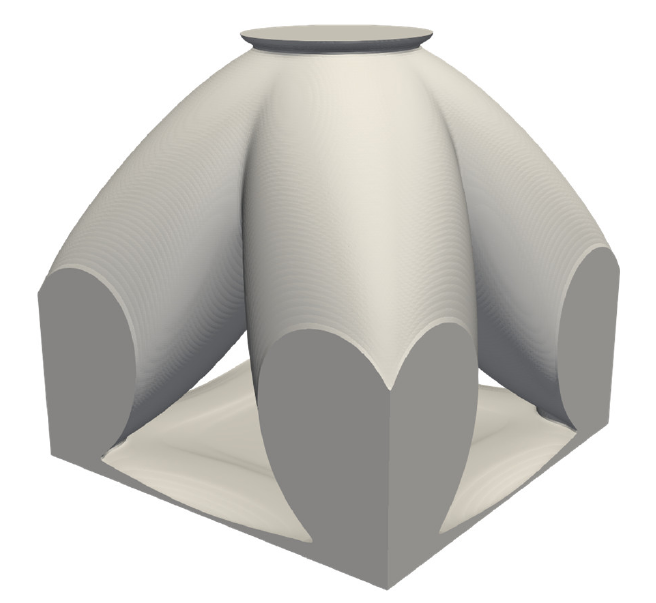}
        \caption{eCNNTO}
        \label{fig:3Dstructurescase42}
    \end{subfigure}

    \vspace{0.5em}
    \caption{Optimized structures in 3D using SIMP and eCNNTO: (a, b) Cantilever beam 1, (c, d) Cantilever beam 2, (e, f) Long beam, and (g, h) Short beam. In each pair, the former is obtained by SIMP and the latter by eCNNTO.}
    \label{fig:all3Doptimizedstructures}
\end{figure*}

The results of acceleration and compliance are summarized in \autoref{table:3D result}. eCNNTO can achieve $76.0\% \sim 96.8\%$ acceleration on these examples. As shown in \autoref{fig:all3Doptimizedstructures}, very similar optimized structures are obtained using SIMP and eCNNTO, which can be further confirmed by the almost identical compliance values in \autoref{table:3D result}. Therefore, strong generalization capabilities are observed also in 3D problems. 3D topology optimization, owing to its substantial number of elements, is prone to exhibiting sluggish convergence; see the long beam and the short beam for examples. In contrast, eCNNTO can learn the density evolution patterns and then bypass the sluggish region, thereby achieving a rapid convergence. 

The significant speedup can be better shown in terms of runtime; see \autoref{table:3D time comparison}. We observe that eCNNTO usually takes less than $20\%$ of the runtime SIMP.

\subsection{Effect of the window size}\label{Sec45}

In this section, we investigate the impact the window size (i.e., $W$ in \autoref{fig:network architecture}) on the acceleration and compliance. We consider window sizes of 3, 5, and 7. All models are trained on the same dataset (see Figures \ref{fig:training benchmarks} and \ref{fig:3D training benchmarks}), where the only difference is the window size. Examples in both 2D (\autoref{fig:2Dcase}) and 3D (\autoref{fig:3Dcase}) are tested.

\begin{table*}
\caption{Results of 2D test problems with different window sizes.}\label{table:2D result dw}%
\centering
\setlength{\tabcolsep}{4pt}
\begin{tabular}{@{}lcccccc@{}}
\toprule
\multirow{2}{*}{Example} & \multicolumn{3}{c}{\# Iterations} & \multicolumn{3}{c}{Compliance}\\
\cmidrule(lr){2-4} \cmidrule(lr){5-7}
& Window 3& Window 5& Window 7 &Window 3 & Window 5 & Window 7\\
\midrule
Long beam    & 79    & \textbf{71}   & 80            & 81.00  & \textbf{80.46}  & 81.26 \\
Square & 305   & \textbf{226}  & 397           & 22.09  & 22.09           & \textbf{22.04} \\
Column        & 94    & \textbf{85}   & 93            & 143.38 & 143.64          & \textbf{143.37}\\
L-shaped      & 268   & \textbf{205}  & 369           & 89.62  & 90.09           & \textbf{89.52} \\
UDL     & 136   & 113           & \textbf{107}  & 1.13   & 1.17            & \textbf{1.11}\\
\bottomrule
\end{tabular}
\end{table*}

\begin{table*}
    \caption{Training cost of 2D models with different window sizes.}\label{table:2D computation cost}
\centering  
\setlength{\tabcolsep}{4pt}
\begin{tabular}{@{}ccc@{}}
\toprule
 Window size & Memory (MB) & Training time per epoch (s) \\
\midrule
          3           & 88.08               & 1.05           \\
          5           & 243.94              & 2.40           \\
          7           & 477.73              & 3.43           \\
\bottomrule
\end{tabular}
\end{table*}
 
The results of 2D examples are listed in \autoref{table:2D result dw}. We observe that the window size affects both the acceleration performance and the compliance. The model with $W=5$ requires the fewest iterations in most examples, whereas $W=7$ yields marginally lower compliance values. However, this minor improvement comes at the cost of a substantial increase in SIMP iterations, which implies that a large window size ($W=7$) may introduce redundant neighborhood information that hinders convergence. Moreover, enlarging the window size from $W=5$ to $W=7$ nearly doubles the memory requirement and increases the training time per epoch by over 40\%; see \autoref{table:2D computation cost}. Since the slight performance gain of $W=7$ does not justify this disproportionate cost in both computational overhead and convergence iterations, a window size of $W=5$ is identified as the most balanced option for accelerating 2D TO problems.

\begin{table*}
\caption{Results of 3D test problems with different window sizes.}\label{table:3D result dw}%
\centering
\setlength{\tabcolsep}{4pt}
\begin{tabular}{@{}lcccccc@{}}
\toprule
\multirow{2}{*}{Example} & \multicolumn{3}{c}{\# Iterations} & \multicolumn{3}{c}{Compliance}\\
\cmidrule(lr){2-4} \cmidrule(lr){5-7}
& Window 3& Window 5& Window 7 &Window 3 & Window 5 & Window 7 \\
\midrule
Cantilever 1   & 195         & 214            & \textbf{189} & 1.075  & 1.070  & \textbf{1.065} \\
Cantilever 1   & 103         & \textbf{77}    & 101          & 0.978  & 0.997  & \textbf{0.976} \\
Long beam      & 90          & 114            & \textbf{74}  & 1.709  & 1.710  & \textbf{1.707} \\
Short beam     & \textbf{65} & 68             & 81           & 3.169  & 3.169  & 3.169 \\
\bottomrule
\end{tabular}
\end{table*}

\begin{table*}
\caption{Training cost of 3D models with different window sizes.}\label{table:3D computation cost}
\centering  
\setlength{\tabcolsep}{4pt}
\begin{tabular}{@{}ccc@{}}
\toprule
Window size & Memory (MB) & Training time per epoch (s) \\
\midrule
          3           & 1239.78             & 18.20          \\
          5           & 5725.86             & 73.51          \\
          7           & 15705.11            & 164.60         \\
\bottomrule
\end{tabular}
\end{table*}

Moreover, the results of 3D examples are summarized in \autoref{table:3D result dw}. We observe that increasing the window size from 3 decreases both the number of iterations and the compliance in most examples. However, while a larger window size ($W=7$) yields marginally lower objective values in all examples, it does not consistently minimize the number of SIMP iterations. Due to the high dimensionality, enlarging the window from $W=3$ to $W=7$ increases the memory requirement by more than 13 times (from approximately 1.2 GB to 15.7 GB) and increases the training time per epoch by more than 9 times; see \autoref{table:3D computation cost}. Since the marginal gains offered by a larger window cannot justify its prohibitive cost in compute, $W=3$ is identified as the best option for 3D topology optimization.

\subsection{Non-design domains}\label{Sec46}

\begin{figure*}
\centering

    \begin{subfigure}{0.36\textwidth}
        \includegraphics[width=\linewidth]{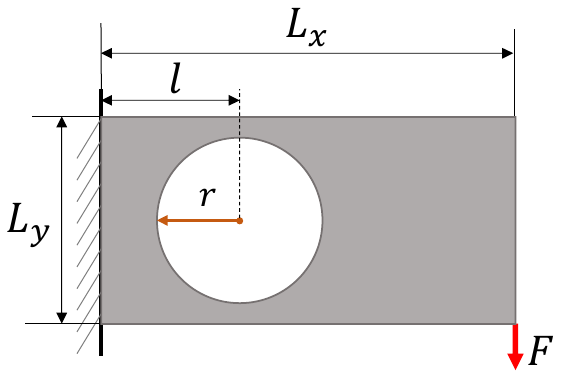}
        \caption{Problem setting}
        \label{fig:Non-design1}
    \end{subfigure}
    \hfill 
    \begin{subfigure}{0.30\textwidth}
        \includegraphics[width=\linewidth]{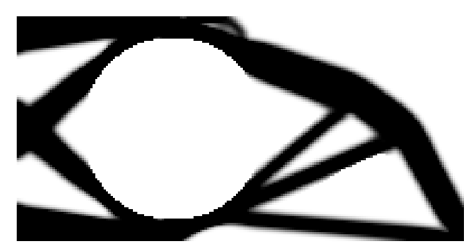}
        \captionsetup{skip=13pt}
        \caption{SIMP}
        \label{fig:Non-design2}
    \end{subfigure}
    \hfill
    \begin{subfigure}{0.30\textwidth}
        \includegraphics[width=\linewidth]{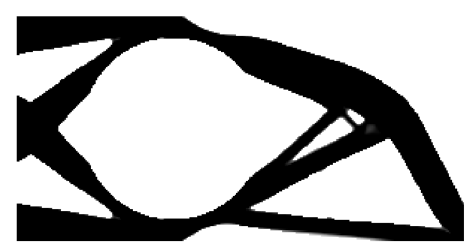}
        \captionsetup{skip=13pt}
        \caption{eCNNTO}
        \label{fig:Non-design3}
    \end{subfigure}

    \vspace{1em}

    \begin{subfigure}{0.36\textwidth}
        \includegraphics[width=\linewidth]{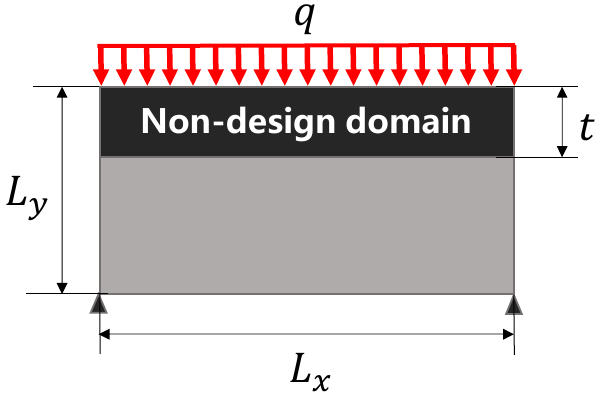}
        \caption{Problem setting}
        \label{fig:Non-design4}
    \end{subfigure}
    \hfill 
    \begin{subfigure}{0.30\textwidth}
        \includegraphics[width=\linewidth]{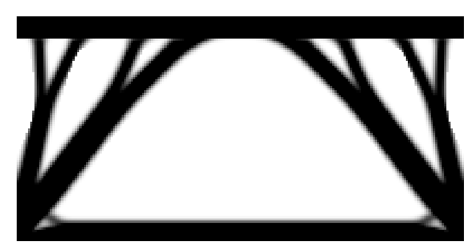}
        \captionsetup{skip=22pt}
        \caption{SIMP}
        \label{fig:Non-design5}
    \end{subfigure}
    \hfill
    \begin{subfigure}{0.30\textwidth}
        \includegraphics[width=\linewidth]{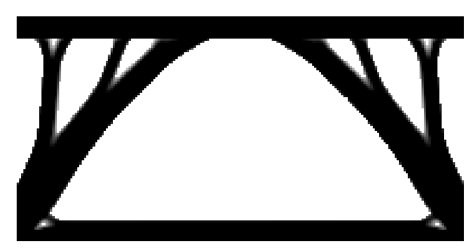}
        \captionsetup{skip=22pt}
        \caption{eCNNTO}
        \label{fig:Non-design6}
    \end{subfigure}

\caption{Topology optimization with non-design domains. (a) A cantilever beam with a hole as the non-design domain, where $L_{x}=2$, $L_{y}=1$, $r=0.4$, $l=0.7$, and $F=1$, (b, c) optimized structures of the cantilever beam by SIMP and eCNNTO, (d) a bridge with a deck on the top as the non-design domain, where $L_{x}=2$, $L_{y}=1$, $t=0.1$, and $q=1$, and (e, f) optimized structures of the bridge by SIMP and eCNNTO.}\label{fig:Non-design}
\end{figure*}

Last but not least, to further demonstrate the generalization capabilities, we apply eCNNTO to TO problems with non-design domains, which are often encountered when certain important structural features are desired to be kept during optimization. Elements in non-design domains are referred to as passive elements, whose densities are prescribed as either void or solid throughout the entire optimization process \citep{andreassenEfficientTopologyOptimization2011}. We evaluate the performance of eCNNTO on two 2D test examples; see \autoref{fig:Non-design}(a, d). Again, the model has been trained and it is directly used here. The first example is a cantilever beam, where a circular hole inside the structure is a desired feature, so it is a non-design domain. The second example is a bridge with a deck as the non-design domain to ensure sufficient load-bearing capacity. Both examples use a $200 \times 200$ mesh. Their volume fractions are 0.5 and 0.4, respectively.

As illustrated in \autoref{fig:Non-design}, the optimized structures of eCNNTO strictly adhere to non-design constraints and exhibit similar features to the those obtained by SIMP. Furthermore, eCNNTO achieves substantial speedups of 79.7\% (cantilever) and 89.0\% (bridge) with even smaller compliance values, as shown in \autoref{table:Non-design result}. Due to the hard constraints imposed on the element densities in non-design domains, the density transition across interfaces between non-design and design domains tends to be discontinuous, which causes SIMP to demand more iterations to enhance continuity and reach convergence. In contrast, when predicting densities at these interfaces, the spatial correlation of eCNNTO already accounts for the density constraints by non-design domains and thus produces smooth transitions. For this reason, it can significantly reduce the number of subsequent SIMP iterations without compromising the optimized compliance value.

\begin{table*}
\caption{Optimization results of non-design domain examples.}\label{table:Non-design result}%
\centering
\setlength{\tabcolsep}{3pt}
\begin{tabular}{lccccc}
\toprule
\multirow{2}{*}{Example} & \multicolumn{2}{c}{\# Iterations} & \multirow{2}{*}{Acceleration (\%)} & \multicolumn{2}{c}{Compliance}\\
\cmidrule(lr){2-3} \cmidrule(lr){5-6}
& SIMP & eCNNTO & & SIMP & eCNNTO\\
\midrule
Cantilever   & 522  & 106  & 79.7 & 152.01  & \textbf{149.43}\\
Bridge       & 881  & 97   & 89.0 & 2.34    & \textbf{2.27} \\
\bottomrule
\end{tabular}
\end{table*}

\section{Conclusion}\label{Sec5}

In this work, we propose eCNNTO, an element-based Convolutional Neural Network (CNN) designed to significantly accelerate density-based topology optimization by learning the local density evolution at the element level. By integrating CNN and residual connections, the network explicitly captures the spatial correlations among neighboring elements, thereby fundamentally enhancing structural connectivity. Data generation is highly efficient in that it only requires to run a few simple benchmark problems. Furthermore, we introduce a novel training strategy by selecting features at the final stage for training, which substantially reduces the training data size. By accounting for spatial dependencies, eCNNTO effectively suppresses isolated structural pieces and intermediate defects, consistently yielding physically valid structures in various cases.

Once the model is trained on simple settings, it can be directly applied to a variety of unseen boundary conditions, loading cases, design domain geometries, mesh resolutions, and non-design domains, consistently demonstrating its strong generalization capabilities and significant speedups. Compared to SIMP, eCNNTO reduces up to 90\% of iterations in 2D and 97\% of iterations in 3D, while maintaining comparable, or in most cases, better compliance values. Compared to DLTOP, eCNNTO requires a significantly smaller data size for training, exhibits much better structural connectivity, and also shows faster speedups.

In the future, several promising directions are worth explorations. For example, applying the proposed model to multi-physics topology optimization may help tackle the curse of dimensionality in such problems. Adoption of graph neural networks (GNNs) may be useful to facilitate the application of eCNNTO to unstructured mesh scenarios. Finally, incorporating volume constraints in the process of learning may be worth investigating as it can further reduce the required SIMP iterations after the eCNNTO prediction.

\section*{Acknowledgments}
S. Lu and X. Wei are partially supported by National Natural Science Foundation of China (No. 12571408 and No. 12494550/12494555).

\appendix

\section{Impact of training strategies}\label{secA1}

\begin{figure*}
\centering

    \begin{subfigure}{0.32\textwidth}
        \includegraphics[width=\linewidth]{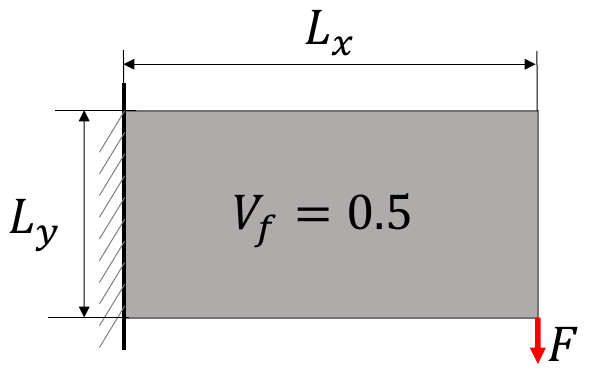}
        \captionsetup{skip=17pt}
        \caption{}
        \label{fig:Appendixcase1}
    \end{subfigure}
    \hfill 
    \begin{subfigure}{0.32\textwidth}
        \includegraphics[width=\linewidth]{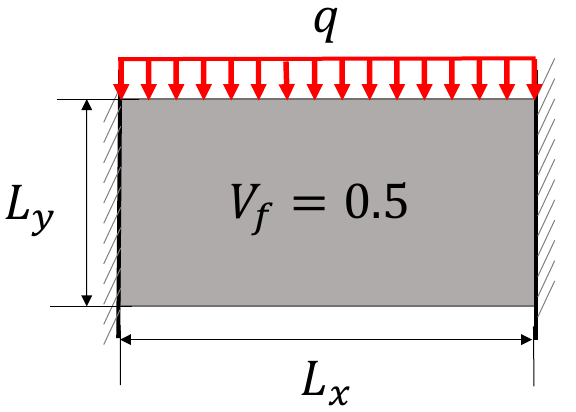}
        \captionsetup{skip=9pt}
        \caption{}
        \label{fig:Appendixcase2}
    \end{subfigure}
    \hfill
    \begin{subfigure}{0.32\textwidth}
        \includegraphics[width=\linewidth]{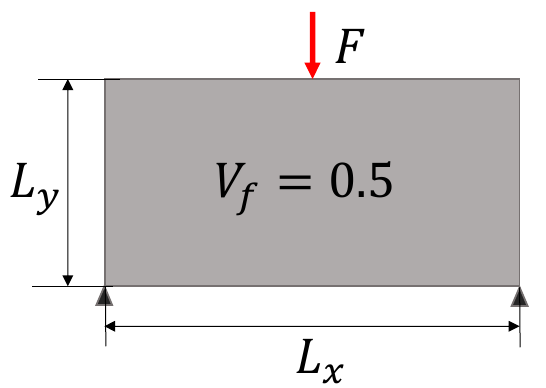}
        \caption{}
        \label{fig:Appendixcase3}
    \end{subfigure}

\caption{Three test examples used to study the impact of different training strategies. (a) A cantilever beam subjected to a concentrated force $F=1$ at its lower-right corner, where $L_{x}=2$ and $L_{y}=1$, (b) a beam clamped on both ends with a uniformly distributed force $q=1$ on the top, where $L_{x}=2$ and $L_{y}=1$, and (c) a simply supported beam subjected to a concentrated force $F=1$ on the top, where $L_{x}=2$ and $L_{y}=\frac{1}{3}$.}\label{fig:Appendixcase}
\end{figure*}

In order not to distract readers from the test problems in the main text, we postpone the study of the impact of different training strategies here. We compare the three different training strategies with the data prepared in \autoref{Sec32}; see \autoref{fig:training benchmarks}. That is, we train three models with features from the Early Stage, the Middle Stage, and the Final Stage, respectively. Once trained, they are tested on the same set of test problems in 2D; see \autoref{fig:Appendixcase}. These test problems use the same settings: the sensitivity filter with radius 3, a mesh of $80 \times 80$, and the number of input size $N=24$.

\begin{figure*}[htbp]
    \centering

    \begin{subfigure}{0.32\textwidth}
        \centering
        \includegraphics[width=\linewidth]{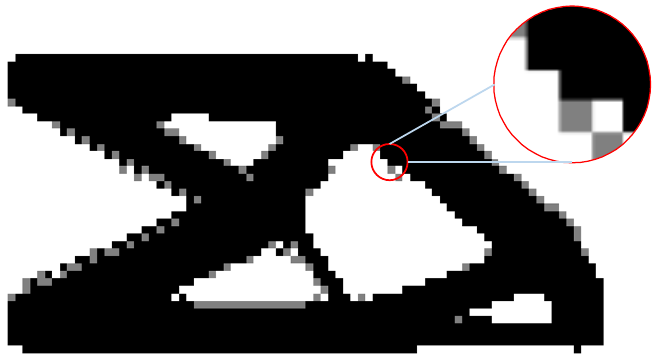}
        \caption{Early Stage}
        \label{fig:Appendixcase1ES}
    \end{subfigure}
    \hfill 
    \begin{subfigure}{0.32\textwidth}
        \centering
        \includegraphics[width=\linewidth]{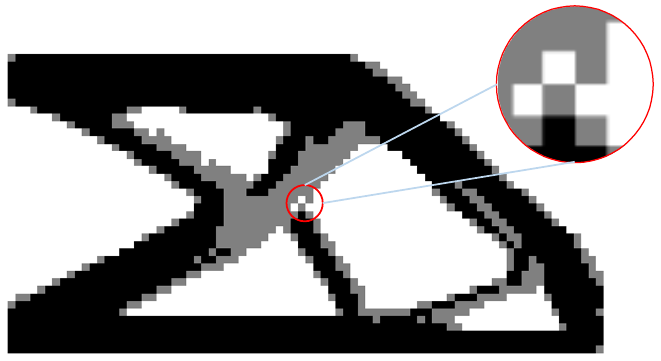}
        \caption{Middle Stage}
        \label{fig:Appendixcase1MS}
    \end{subfigure}
    \hfill
    \begin{subfigure}{0.32\textwidth}
        \centering
        \includegraphics[width=\linewidth]{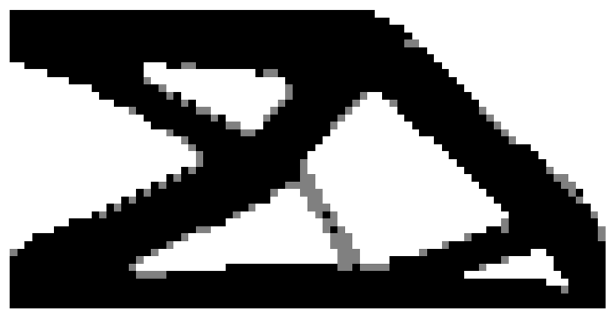}
        \caption{Final Stage}
        \label{fig:Appendixcase1FS}
    \end{subfigure}
    
    \vspace{1.5em} 
    
    \begin{subfigure}{0.32\textwidth}
        \centering
        \includegraphics[width=\linewidth]{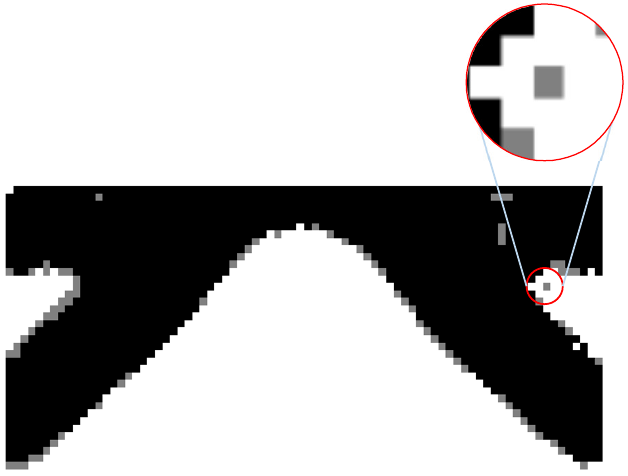}
        \captionsetup{skip=10pt}
        \caption{Early Stage}
        \label{fig:Appendixcase2ES}
    \end{subfigure}
    \hfill 
    \begin{subfigure}{0.32\textwidth}
        \centering
        \includegraphics[width=\linewidth]{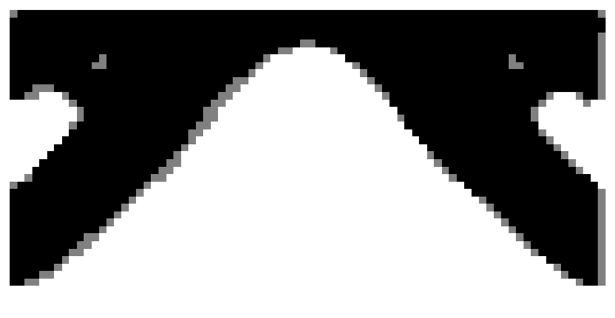}
        \caption{Middle Stage}
        \label{fig:Appendixcase2MS}
    \end{subfigure}
    \hfill
    \begin{subfigure}{0.32\textwidth}
        \centering
        \includegraphics[width=\linewidth]{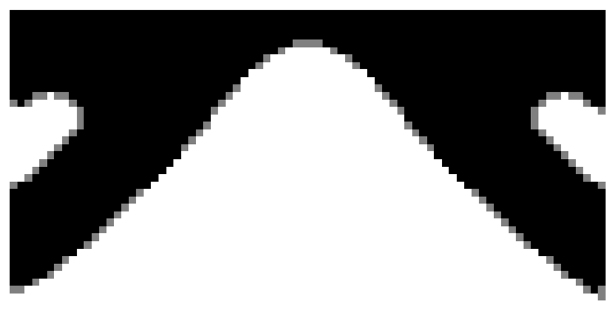}
        \caption{Final Stage}
        \label{fig:Appendixcase2FS}
    \end{subfigure}
    
    \vspace{1.5em} 
    
    \begin{subfigure}{0.32\textwidth}
        \centering
        \includegraphics[width=\linewidth]{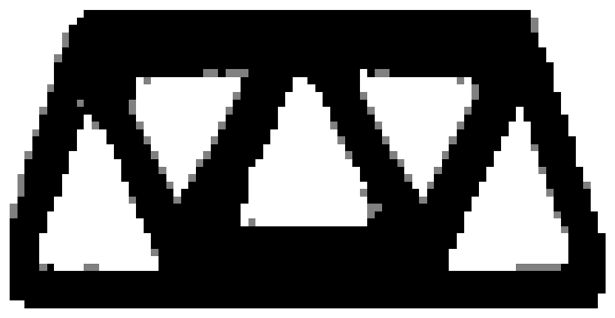}
        \caption{Early Stage}
        \label{fig:Appendixcase3ES}
    \end{subfigure}
    \hfill 
    \begin{subfigure}{0.32\textwidth}
        \centering
        \includegraphics[width=\linewidth]{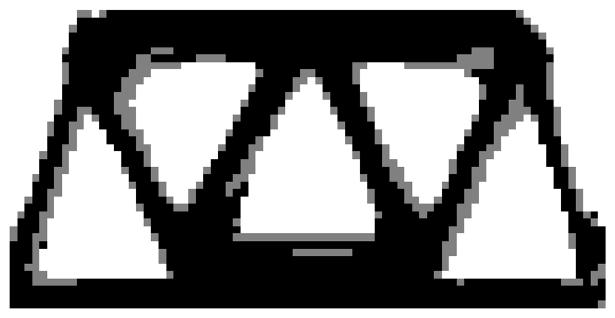}
        \caption{Middle Stage}
        \label{fig:Appendixcase3MS}
    \end{subfigure}
    \hfill
    \begin{subfigure}{0.32\textwidth}
        \centering
        \includegraphics[width=\linewidth]{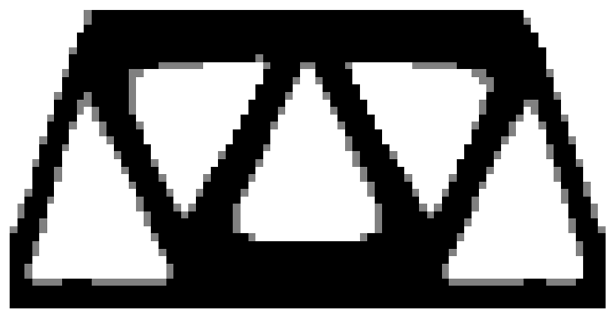}
        \caption{Final Stage}
        \label{fig:Appendixcase3FS}
    \end{subfigure}

    \vspace{1em}
    \caption{Comparison of predicted structures using models trained with different features. Each row corresponds to a test problem in \autoref{fig:Appendixcase}, whereas each column corresponds to a particular training strategy.}
    \label{fig:Appendixallpredictedcases}
\end{figure*}

The predicted structures by these models are shown in \autoref{fig:Appendixallpredictedcases}. Focusing on the predictions of the Early-Stage model, we observe that jagged boundaries appear quite often. This arises from the noise in the early density evolution, which compromises learning outcomes and prevents smooth structural transitions. The predictions of the Middle-Stage model, they exhibit more intermediate-density regions around the boundaries. This indicates that the model fails to make correct predictions for boundary elements, resulting in numerous intermediate-density regions in the predicted structures and even longer iteration counts than SIMP. In contrast, the model trained on the Final Stage features produces valid structures with clear boundaries for all the three examples while achieving significant speedup.

\newpage
\bibliographystyle{elsarticle-harv} 
\bibliography{bibliography}

\end{document}